\documentclass[twoside,11pt]{article}
\usepackage{graphicx}
\usepackage{graphics}
\usepackage{multirow}
\usepackage{subfigure}
\usepackage{xspace}
\usepackage{jair}
\usepackage{theapa}
\usepackage{url}

\jairheading{27}{2006}{119--151}{1/06}{10/06} \ShortHeadings{A
Comparison of Machine Transliteration Models} {Oh, Choi, \& Isahara}
\firstpageno{119}

\begin{document}

\title{A Comparison of Different Machine Transliteration Models}

\author{\name Jong-Hoon Oh \email rovellia@nict.go.jp \\
       \addr Computational Linguistics Group\\
       National Institute of Information and Communications
       Technology (NICT)\\
       3-5 Hikaridai, Seika-cho, Soraku-gun, Kyoto 619-0289
       Japan\\
        \AND
       \name Key-Sun Choi \email kschoi@cs.kaist.ac.kr \\
       \addr Computer Science Division, Department of EECS \\
       Korea Advanced Institute of Science and Technology (KAIST)\\
        373-1 Guseong-dong, Yuseong-gu, Daejeon 305-701 Republic of Korea \\
       \AND
       \name Hitoshi Isahara \email isahara@nict.go.jp \\
       \addr Computational Linguistics Group\\
       National Institute of Information and Communications
       Technology (NICT)\\
       3-5 Hikaridai, Seika-cho, Soraku-gun, Kyoto 619-0289
       Japan
       }


\maketitle

\begin{abstract}
Machine transliteration is a method for automatically converting
words in one language into phonetically equivalent ones in another
language. Machine transliteration plays an important role in
natural language applications such as information retrieval and
machine translation, especially for handling proper nouns and
technical terms. Four machine transliteration models --
grapheme-based transliteration model, phoneme-based
transliteration model, hybrid transliteration model, and
correspondence-based transliteration model -- have been proposed
by several researchers. To date, however, there has been little
research on a framework in which multiple transliteration models
can operate simultaneously. Furthermore, there has been no
comparison of the four models within the same framework and using
the same data. We addressed these problems by 1) modeling the four
models within the same framework, 2) comparing them under the same
conditions, and 3) developing a way to improve machine
transliteration through this comparison. Our comparison showed
that the hybrid and correspondence-based models were the most
effective and that the four models can be used in a complementary
manner to improve machine transliteration performance.
\end{abstract}

\section{Introduction}
\label{sec:introduction}
    With the advent of new technology and the flood of information
through the Web, it has become increasingly common to adopt foreign
words into one's language.
    This usually entails adjusting
the adopted word's original pronunciation to follow the phonological
rules of the target language, along with modification of its
orthographical form.
    This phonetic ``translation" of foreign words is called \textit{transliteration}.
    For example, the English word \textit{data} is transliterated into Korean as
`de-i-teo'\footnote{In this paper, target language transliterations
are represented in their Romanized form with single quotation marks
and hyphens between syllables.} and into Japanese as `de-e-ta'.
    Transliteration is particularly used to
translate proper names and technical terms from languages using
Roman alphabets into ones using non-Roman alphabets such as from
English to Korean, Japanese, or Chinese.
    Because transliteration
is one of the main causes of the out-of-vocabulary (OOV) problem,
transliteration by means of dictionary lookup is
impractical~\cite{fujii01,lin02}.
    One way to solve the OOV problem is to use machine transliteration.
    Machine transliteration is usually used to support machine translation
(MT)~\cite{knight97,al-onaizan02} and cross-language information
retrieval (CLIR)~\cite{fujii01,lin02}.
    For CLIR, machine transliteration bridges the gap between the transliterated
localized form and its original form by generating all possible
transliterations from the original form (or generating all possible
original forms from the transliteration)\footnote{The former process
is generally called ``transliteration", and the latter is generally
called ``back-transliteration"~\cite{knight97}}.
    For example, machine transliteration can assist query translation in CLIR,
where proper names and technical terms frequently appear in source
language queries.
    In the area of MT, machine transliteration
helps preventing translation errors when translations of proper
names and technical terms are not registered in the translation
dictionary.
    Machine transliteration can therefore improve the
performance of MT and CLIR.

    Four machine transliteration models have been proposed by several
researchers: \textbf{graph-eme}\footnote{\textit{Graphemes} refer to
the basic units (or the smallest contrastive units) of a written
language: for example, English has 26 graphemes or letters, Korean
has 24, and German has 30.}\textbf{-based transliteration model}
($\psi_G$)~\cite{lee98,jeong99,kim99,lee99,kangbj00,kangih00,kang01,goto03,li04},
\textbf{phoneme}\footnote{\textit{Phonemes} are the simplest
significant unit of sound (or the smallest contrastive units of a
spoken language); for example, /M/, /AE/, and /TH/ in /M AE TH/, the
pronunciation of \textit{math}. We use the \textit{ARPAbet} symbols
to represent source phonemes. \textit{ARPAbet} is one of the methods
used for coding source phonemes into ASCII characters
(\url{http://www.cs.cmu.edu/~laura/pages/arpabet.ps}). Here we
denote source phonemes and pronunciation with two slashes, as in
/AH/, and use pronunciation based on \textit{The CMU Pronunciation
Dictionary} and \textit{The American Heritage(r) Dictionary of the
English Language}.}\textbf{-based transliteration model}
($\psi_P$)~\cite{knight97,lee99,jung00,meng01}, \textbf{hybrid
transliteration model}
($\psi_{H}$)~\cite{lee99,al-onaizan02,bilac04}, and
\textbf{correspondence-based transliteration model}
($\psi_{C}$)~\cite{oh02}.
    These models are classified in terms of the units to be
transliterated.
    The $\psi_G$ is sometimes referred to as the
\textit{direct method} because it directly transforms source
language graphemes into target language graphemes without any
phonetic knowledge of the source language words.
    The $\psi_P$ is sometimes referred to as the \textit{pivot method} because it
uses source language phonemes as a pivot when it produces target
language graphemes from source language graphemes.
    The $\psi_P$ therefore usually needs two steps: 1) produce
source language phonemes from source language graphemes; 2) produce
target language graphemes from source phonemes\footnote{ These two
steps are explicit if the transliteration system produces target
language transliterations after producing the pronunciations of the
source language words; they are implicit if the system uses phonemes
implicitly in the transliteration stage and explicitly in the
learning stage, as described elsewhere~\cite{bilac04}}.
    The $\psi_{H}$ and $\psi_{C}$
make use of both source language graphemes and source language
phonemes when producing target language transliterations. Hereafter,
we refer to a source language grapheme as a \textit{source
grapheme}, a source language phoneme as a \textit{source phoneme},
and a target language grapheme as a \textit{target grapheme}.

    The transliterations produced by the four models usually differ
because the models use different information.
    Generally, transliteration is a phonetic process, as in $\psi_P$, rather than
an orthographic one, as in $\psi_G$~\cite{knight97}.
    However, standard transliterations are not restricted to phoneme-based
transliterations.
    For example, the standard Korean transliterations of \textit{data},
\textit{amylase}, and \textit{neomycin} are, respectively, the
phoneme-based transliteration `de-i-teo', the grapheme-based
transliteration `a-mil-la-a-je', and `ne-o-ma-i-sin', which is a
combination of the grapheme-based transliteration `ne-o' and the
phoneme-based transliteration `ma-i-sin'.
    Furthermore, if the unit to be transliterated is restricted to
either a source grapheme or a source phoneme, it is hard to produce
the correct transliteration in many cases.
    For example, $\psi_P$ cannot easily produce the grapheme-based transliteration
`a-mil-la-a-je', the standard Korean transliteration of
\textit{amylase}, because $\psi_P$ tends to produce `a-mil-le-i-seu'
based on the sequence of source phonemes /AE M AH L EY S/.
    Multiple transliteration models should therefore be applied to
better cover the various transliteration processes.
    To date, however, there has been little published research regarding a
framework in which multiple transliteration models can operate
simultaneously.
    Furthermore, there has been no reported comparison
of the transliteration models within the same framework and using
the same data although many English-to-Korean transliteration
methods based on $\psi_G$ have been compared to each other with the
same data~\cite{kangbj00,kangih00,oh02}.

    To address these problems, we 1) \textbf{modeled a framework
in which the four transliteration models can operate
simultaneously}, 2) \textbf{compared the transliteration models
under the same conditions}, and 3) \textbf{using the results of the
comparison, developed a way to improve the performance of machine
transliteration}.

    The rest of this paper is organized as follows.
    Section~\ref{sec:related} describes previous work relevant to our
study.
    Section~\ref{sec:mt} describes our implementation of the
four transliteration models.
    Section~\ref{sec:exp} describes our testing and results.
    Section~\ref{sec:discuss} describes a way to improve
machine transliteration based on the results of our comparison.
    Section~\ref{sec:ranking} describes a transliteration ranking
method that can be used to improve transliteration performance.
    Section~\ref{sec:conclusion} concludes the paper with a summary
and a look at future work.


\section{Related Work}
\label{sec:related}
    Machine transliteration has received significant research
attention in recent years.
    In most cases, the source language and
target language have been English and an Asian language,
respectively -- for example, English to Japanese~\cite{goto03},
English to Chinese~\cite{meng01,li04}, and English to
Korean~\cite{lee98,kim99,jeong99,lee99,jung00,kangbj00,kangih00,kang01,oh02}.
    In this section, we review previous work related to the four
transliteration models.

\subsection{Grapheme-based Transliteration Model}
\label{subsec:gt}

    Conceptually, the $\psi_G$ is direct orthographical mapping
from source graphemes to target graphemes.
    Several transliteration
methods based on this model have been proposed, such as those based
on a source-channel model~\cite{lee98,lee99,jeong99,kim99}, a
decision tree~\cite{kangbj00,kang01}, a transliteration
network~\cite{kangih00,goto03}, and a joint source-channel
model~\cite{li04}.

    The methods based on the source-channel model
deal with English-Korean transliteration.
    They use a chunk of graphemes that can correspond to a source phoneme.
    First, English words are segmented into a chunk of English graphemes.
    Next, all possible chunks of Korean graphemes corresponding to the
chunk of English graphemes are produced.
    Finally, the most relevant sequence of Korean graphemes is identified by using the
source-channel model.
    The advantage of this approach is that it
considers a chunk of graphemes representing a phonetic property of
the source language word.
    However, errors in the first step (segmenting the English words) propagate to
the subsequent steps, making it difficult to produce correct
transliterations in those steps. Moreover, there is high time
complexity because all possible chunks of graphemes are generated in
both languages.

    In the method based on a decision tree, decision trees that
transform each source grapheme into target graphemes are learned and
then directly applied to machine transliteration. The advantage of
this approach is that it considers a wide range of contextual
information, say, the left three and right three contexts.
    However, it does not consider any phonetic aspects of
transliteration.

    Kang and Kim~\citeyear{kangih00} and Goto~\textit{et
al.}~\citeyear{goto03} proposed methods based on a transliteration
network for, respectively, English-to-Korean and English-to-Japanese
transliteration.
    Their frameworks for constructing a transliteration network are similar -- both are
composed of nodes and arcs.
    A node represents a chunk of source
graphemes and its corresponding target graphemes.
    An arc represents a possible link between nodes and has a weight showing its
strength.
    Like the methods based on the source-channel model,
their methods consider the phonetic aspect in the form of chunks of
graphemes.
    Furthermore, they segment a chunk of graphemes and
identify the most relevant sequence of target graphemes in one step.
    This means that errors are not propagated from one step to
the next, as in the methods based on the source-channel model.

    The method based on the joint source-channel model simultaneously
considers the source language and target language contexts (bigram
and trigram) for machine transliteration.
    Its main advantage is the use of bilingual contexts.

\subsection{Phoneme-based Transliteration Model}
\label{subsec:pt}

    In the $\psi_P$, the transliteration
key is pronunciation or the source phoneme rather than spelling or
the source grapheme.
    This model is basically \textit{source grapheme-to-source phoneme}
transformation and \textit{source phoneme-to-target grapheme}
transformation.

    Knight and Graehl~\citeyear{knight97} modeled Japanese-to-English
transliteration with weighted finite state transducers (WFSTs) by
combining several parameters including romaji-to-phoneme,
phoneme-to-English, English word probabilities, and so on.
    A similar model was developed for Arabic-to-English
transliteration~\cite{stalls98}.
    Meng \textit{et al.}~\citeyear{meng01} proposed an English-to-Chinese
transliteration method based on English grapheme-to-phoneme
conversion, cross-lingual phonological rules, mapping rules between
English phonemes and Chinese phonemes, and Chinese syllable-based
and character-based language models.
    Jung \textit{et al.}~\citeyear{jung00} modeled English-to-Korean
transliteration with an extended Markov window.
    The method transforms an English word into English pronunciation by using a
pronunciation dictionary.
    Then it segments the English phonemes into chunks of English phonemes;
each chunk corresponds to a Korean grapheme as defined by
handcrafted rules.
    Finally, it automatically transforms each chunk
of English phonemes into Korean graphemes by using an extended
Markov window.

    Lee~\citeyear{lee99} modeled English-to-Korean transliteration in
two steps.
    The \textit{English grapheme-to-English phoneme}
transformation is modeled in a manner similar to his method based on
the source-channel model described in Section~\ref{subsec:gt}.
    The English phonemes are then transformed into Korean graphemes by
using English-to-Korean standard conversion rules
(EKSCR)~\cite{min95}.
    These rules are in the form of context-sensitive rewrite rules,
``$P_AP_XP_B \rightarrow y$", meaning that English phoneme $P_X$ is
rewritten as Korean grapheme $y$ in the context $P_A$ and $P_B$,
where $P_X$, $P_A$, and $P_B$ represent English phonemes.
    For example, ``$P_A=\ast,P_X=/SH/,P_B=end \rightarrow$ `si'" means ``English
phoneme $/SH/$ is rewritten into Korean grapheme `si' if it occurs
at the end of the word ($end$) after any phoneme ($\ast$)".
    This approach suffers from both the propagation of errors and the
limitations of EKSCR.
    The first step, grapheme-to-phoneme transformation,
usually results in errors, and the errors propagate to the next
step.
    Propagated errors make it difficult for a transliteration
system to work correctly.
    In addition, EKSCR does not contain
enough rules to generate correct Korean transliterations since its
main focus is mapping from an English phoneme to Korean graphemes
without taking into account the contexts of the English grapheme.

\subsection{Hybrid and Correspondence-based Transliteration Models}
\label{subsec:gpt}

    Attempts to use both source graphemes and source phonemes in
machine transliteration led to the correspondence-based
transliteration model ($\psi_{C}$)~\cite{oh02} and the hybrid
transliteration model
($\psi_{H}$)~\cite{lee99,al-onaizan02,bilac04}.
    The former makes use of the correspondence between a source grapheme and a source
phoneme when it produces target language graphemes; the latter
simply combines $\psi_{G}$ and $\psi_{P}$ through linear
interpolation.
    Note that the $\psi_{H}$ combines the
grapheme-based transliteration probability ($Pr(\psi_{G})$) and the
phoneme-based transliteration probability ($Pr(\psi_{P})$) using
linear interpolation.

    Oh and Choi~\citeyear{oh02} considered the contexts of a source
grapheme and its corresponding source phoneme for English-to-Korean
transliteration.
    They used EKSCR as the basic rules in their method.
    Additional contextual rules are
semi-automatically constructed by examining the cases in which EKSCR
produced incorrect transliterations because of a lack of contexts.
    These contextual rules are in the form of
context-sensitive rewrite rules, ``$C_AC_XC_B \rightarrow y$",
meaning ``$C_X$ is rewritten as target grapheme $y$ in the context
$C_A$ and $C_B$".
    Note that $C_X$, $C_A$, and $C_B$ represent
the correspondence between the English grapheme and phoneme.
    For example, we can read ``$C_A=(\ast: /Vowel/),C_X=(r: /R/),C_B=(\ast:
/Consonant/) \rightarrow$ NULL" as ``English grapheme $r$
corresponding to phoneme $/R/$ is rewritten into null Korean
graphemes when it occurs after vowel phonemes, ($\ast:/Vowel/$),
before consonant phonemes, ($\ast:/Consonant/$)".
    The main advantage of this approach is the application of
a sophisticated rule that reflects the context of the source
grapheme and source phoneme by considering their correspondence.
    However, there is lack of portability to other languages because
the rules are restricted to Korean.

    Several researchers~\cite{lee99,al-onaizan02,bilac04} have
proposed hybrid model-based transliteration methods.
    They model $\psi_{G}$ and $\psi_{P}$ with WFSTs or a source-channel model and
combine $\psi_{G}$ and $\psi_{P}$ through linear interpolation.
    In their $\psi_{P}$, several parameters are
considered, such as the \textit{source grapheme-to-source phoneme}
probability, \textit{source phoneme-to-target grapheme} probability,
and \textit{target language word} probability.
    In their $\psi_{G}$, the \textit{source grapheme-to-target grapheme}
probability is mainly considered.
    The main disadvantage of the hybrid model is that
the dependence between the source grapheme and source phoneme is not
taken into consideration in the combining process; in contrast, Oh
and Choi's approach~\cite{oh02} considers this dependence by using
the correspondence between the source grapheme and phoneme.


\section{Modeling Machine Transliteration Models}
\label{sec:mt}
    In this section, we describe our implementation of the four
machine transliteration models ($\psi_{G}$, $\psi_{P}$, $\psi_{H}$,
and $\psi_{C}$) using three machine learning algorithms:
memory-based learning, decision-tree learning, and the maximum
entropy model.

\subsection{Framework for Four Machine Transliteration Models}

\begin{figure}[ht]
\begin{center}
\includegraphics[width=0.600\textwidth]{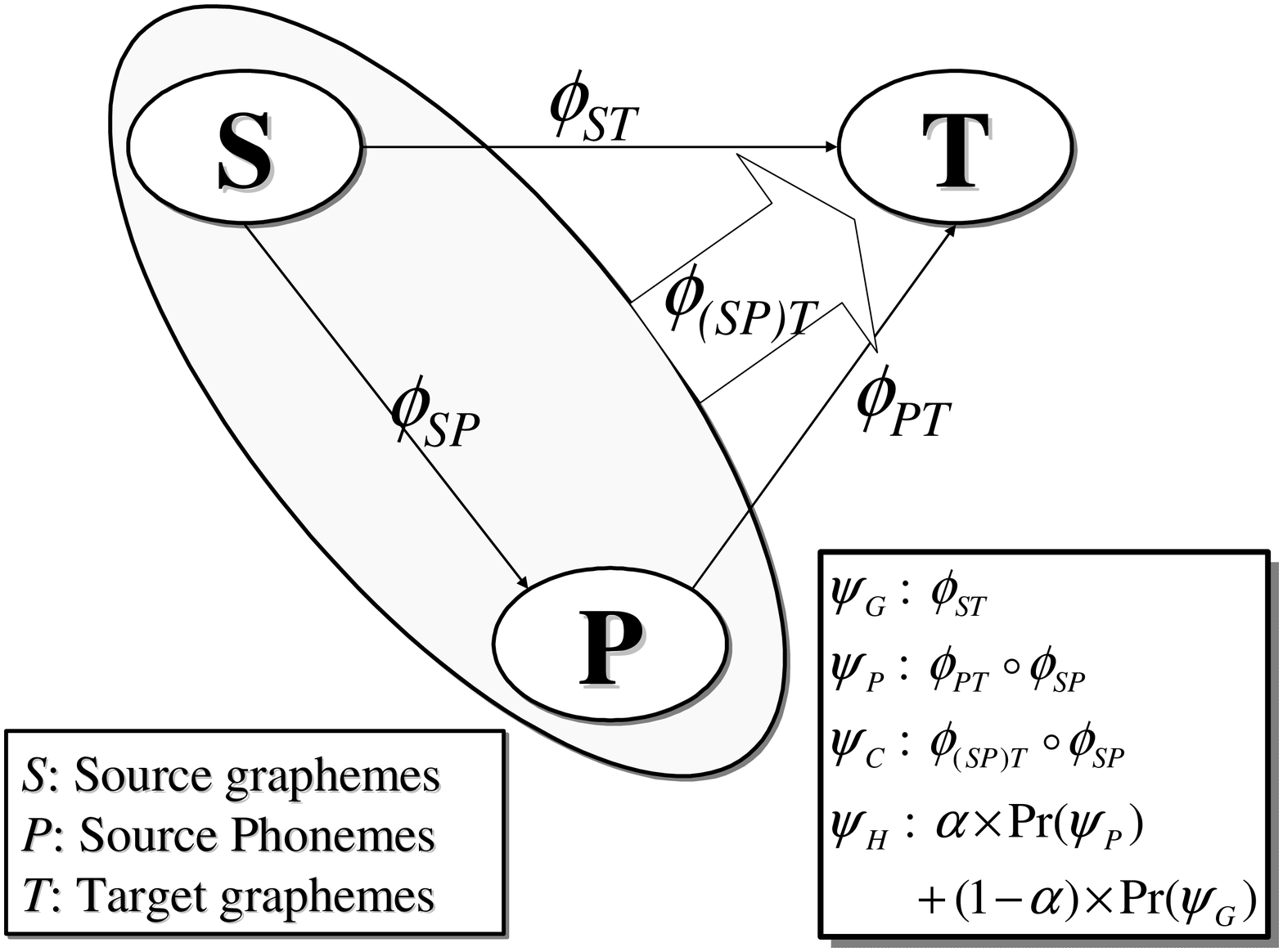}
\end{center}
\caption{Graphical representation of each component function and
four transliteration models: S is a set of source graphemes (e.g.,
letters of the English alphabet), P is a set of source phonemes
defined in \textit{ARPAbet}, and T is a set of target graphemes.}
\label{fig2}
\end{figure}

    Figure~\ref{fig2} summarizes the differences among the
transliteration models and their component functions.
    The $\psi_{G}$ directly transforms source graphemes (S) into
target graphemes (T).
    The $\psi_{P}$ and $\psi_{C}$
transform source graphemes into source phonemes and then generate
target graphemes\footnote{According to $(g \circ f)(x) = g(f(x))$,
we can write $(\phi_{(SP)T} \circ \phi_{SP})(x) =
\phi_{(SP)T}(\phi_{SP}(x))$ and $(\phi_{PT} \circ \phi_{SP})(x) =
\phi_{PT}(\phi_{SP}(x))$.}.
    While $\psi_{P}$ uses only the source phonemes,
$\psi_{C}$ uses the correspondence between the source grapheme and
the source phoneme when it generates target graphemes.
    We describe their differences with two functions, $\phi_{PT}$ and
$\phi_{(SP)T}$.
    The $\psi_{H}$ is represented as the linear
interpolation of $Pr(\psi_{G})$ and $Pr(\psi_{P})$ by means of
$\alpha$ ($0 \leq \alpha \leq 1$).
    Here, $Pr(\psi_{P})$ is the probability that $\psi_{P}$ will produce target graphemes,
while $Pr(\psi_{G})$ is the probability that $\psi_{G}$ will produce
target graphemes.
    We can thus regard $\psi_{H}$ as being
composed of component functions of $\psi_{G}$ and $\psi_{P}$
($\phi_{SP}$, $\phi_{PT}$, and $\phi_{ST}$).
    Here we use the maximum entropy model as the machine learning algorithm for
$\psi_{H}$ because $\psi_{H}$ requires $Pr(\psi_{P})$ and
$Pr(\psi_{G})$, and only the maximum entropy model among
memory-based learning, decision-tree learning, and the maximum
entropy model can produce the probabilities.

    To train each component function, we need to define the features
that represent training instances and data.
    Table~\ref{tab1} shows five feature types, $f_S$, $f_P$, $f_{Stype}$,
$f_{Ptype}$, and $f_T$.
    The feature types used depend on the component functions.
    The modeling of each component function with the feature types is
explained in Sections~\ref{subsec:component} and \ref{subsec:ml}.

\begin{table}[ht]

 \begin{center}
\begin{tabular}{|c|l|l|}
 \hline
 \multicolumn{2}{|c|}{Feature type} & Description and possible values \\
 \hline \hline
 \multirow{4}{*}{$f_{S,Stype}$} &\multirow{2}{*}{$f_{S}$} &Source graphemes in S:\\
 & &26 letters in English alphabet \\
 \cline{2-3}
 & \multirow{2}{*}{$f_{Stype}$} &Source grapheme types: \\
 & & Consonant (C) and Vowel (V) \\ \hline
 \multirow{4}{*}{$f_{P,Ptype}$} & \multirow{2}{*}{$f_P$} & Source phonemes in P \\
 & & (/AA/, /AE/, and so on) \\ \cline{2-3}
 & \multirow{2}{*}{$f_{Ptype}$} & Source phoneme types: Consonant (C), Vowel (V), \\
 & & Semi-vowel (SV), and silence (/$\sim$/) \\ \hline
\multicolumn{2}{|c|}{$f_T$} & Target graphemes in T \\ \hline
\end{tabular}
 \caption{\label{tab1} Feature types used for transliteration models:
 $f_{S,Stype}$ indicates both $f_{S}$ and $f_{Stype}$,
 while $f_{P,Ptype}$ indicates both $f_{P}$ and $f_{Ptype}$.}
 \end{center}
\end{table}

\subsection{Component Functions of Each Transliteration Model}
\label{subsec:component}

\begin{table}[ht]
\begin{center}
\begin{tabular}{|l|l|l|l|}
  \hline
Notation & Feature types used  & Input & Output \\ \hline \hline
    $\phi_{SP}$ & $f_{S,Stype}$, $f_P$ & $s_i, c(s_i)$ & $p_i$
    \\ \hline
    $\phi_{(SP)T}$ & $f_{S,Stype}$, $f_{P,Ptype}$, $f_T$  & $s_i, p_i,c(s_i), c(p_i)$ & $t_i$ \\ \hline
    $\phi_{PT}$ & $f_{P,Ptype}$, $f_T$ & $p_i, c(p_i)$ & $t_i$  \\ \hline
    $\phi_{ST}$ & $f_{S,Stype}$, $f_T$ & $s_i, c(s_i)$ & $t_i$  \\ \hline
\end{tabular}
 \caption{\label{tab2} Definition of each component function:
 $s_i$, $c(s_i)$, $p_i$, $c(p_i)$, and $t_i$ respectively
 represent the $i^{th}$ source grapheme,
 the context of $s_i$ ($s_{i-n},\cdots,s_{i-1}$ and $s_{i+1},\cdots,s_{i+n}$),
 the $i^{th}$ source phoneme,
 the context of $p_i$ ($p_{i-n},\cdots,p_{i-1}$ and $p_{i+1},\cdots,p_{i+n}$),
 and the $i^{th}$ target grapheme. }
 \end{center}
\end{table}

    Table~\ref{tab2} shows the definitions of the four component
functions that we used.
    Each is defined in terms of its input and
output: the first and last characters in the notation of each
correspond respectively to its input and output.
    The role of each component function in each transliteration model is to produce the
most relevant output from its input.
    The performance of a transliteration model therefore depends strongly on that of its
component functions.
    In other words, the better the modeling of
each component function, the better the performance of the machine
transliteration system.

    The modeling strongly depends on the feature type.
    Different feature types are used by the $\phi_{(SP)T}$, $\phi_{PT}$, and
$\phi_{ST}$ functions, as shown in Table~\ref{tab2}.
    These three component functions thus have different strengths and weaknesses
for machine transliteration.
    The $\phi_{ST}$ function is good at
producing grapheme-based transliterations and poor at producing
phoneme-based ones.
    In contrast, the $\phi_{PT}$ function is good
at producing phoneme-based transliterations and poor at producing
grapheme-based ones.
    For \textit{amylase} and its standard Korean
transliteration, `a-mil-la-a-je', which is a grapheme-based
transliteration, $\phi_{ST}$ tends to produce the correct
transliteration; $\phi_{PT}$ tends to produce wrong ones like
`ae-meol-le-i-seu', which is derived from /AE M AH L EY S/, the
pronunciation of \textit{amylase}.
    In contrast, $\phi_{PT}$ can produce `de-i-teo',
which is the standard Korean transliteration of \textit{data} and a
phoneme-based transliteration, while $\phi_{ST}$ tends to give a
wrong one, like `da-ta'.

    The $\phi_{(SP)T}$ function combines the advantages of $\phi_{ST}$
and $\phi_{PT}$ by utilizing the correspondence between the source
grapheme and source phoneme.
    This correspondence enables $\phi_{(SP)T}$ to produce
both grapheme-based and phoneme-based transliterations.
    Furthermore, the correspondence provides important clues for use
in resolving transliteration ambiguities\footnote{Though contextual
information can also be used to reduce ambiguities, we limit our
discussion here to the feature type.}.
    For example, the
source phoneme /AH/ produces much ambiguity in machine
transliteration because it can be mapped to almost every vowel in
the source and target languages (the underlined graphemes in the
following example corresponds to /AH/: hol\underline{o}caust in
English, `hol-l\underline{o}-ko-seu-teu' in its Korean counterpart,
and `ho-r\underline{o}-ko-o-su-to' in its Japanese counterpart).
    If we know the correspondence between the source
grapheme and source phoneme, we can more easily infer the correct
transliteration of /AH/ because the correct target grapheme
corresponding to /AH/ usually depends on the source grapheme
corresponding to /AH/.
    Moreover, there are various Korean
transliterations of the source grapheme \textit{a}: `a', `ae', `ei',
`i', and `o'.
    In this case, the English phonemes
corresponding to the English grapheme can help a component function
resolve transliteration ambiguities, as shown in Table~\ref{tab3}.
    In Table~\ref{tab3}, the \textit{a} underlined in the example
words shown in the last column is pronounced as the English phoneme
in the second column.
    By looking at English grapheme and its corresponding English phoneme,
we can find correct Korean transliterations more easily.

\begin{table}[ht]
\begin{center}
\begin{tabular}{|c|c|l|l|}
  \hline
Korean Grapheme & English Phoneme &Example usage\\ \hline \hline
 `a' &/AA/ & ad\underline{a}gio, saf\underline{a}ri, viv\underline{a}ce \\ \hline
 `ae' &/AE/ & adv\underline{a}ntage, \underline{a}l\underline{a}baster,
tr\underline{a}vertine \\ \hline
 `ei' &/EY/ & ch\underline{a}mber, champ\underline{a}gne,
ch\underline{a}os \\ \hline
 `i' &/IH/ &advant\underline{a}ge, aver\underline{a}ge, sil\underline{a}ge \\ \hline
 `o' & /AO/ & \underline{a}llspice, b\underline{a}ll,
 ch\underline{a}lk \\ \hline
\end{tabular}
 \caption{\label{tab3} Examples of Korean graphemes
derived from English grapheme \textit{a} and its corresponding
English phonemes: the underlines in the example words indicate the
English grapheme corresponding to English phonemes in the second
column.}
\end{center}
\end{table}

    Though $\phi_{(SP)T}$ is more effective than both $\phi_{ST}$ and
$\phi_{PT}$ in many cases, $\phi_{(SP)T}$ sometimes works poorly
when the standard transliteration is strongly biased to either
grapheme-based or phoneme-based transliteration.
    In such cases, either the source grapheme or source phoneme does not contribute
to the correct transliteration, making it difficult for
$\phi_{(SP)T}$ to produce the correct transliteration.
    Because $\phi_{ST}$, $\phi_{PT}$, and $\phi_{(SP)T}$ are the core parts of
$\psi_{G}$, $\psi_{P}$, and $\psi_{C}$, respectively, the advantages
and disadvantages of the three component functions correspond to
those of the transliteration models in which each is used.

    Transliteration usually depends on context.
    For example, the English grapheme \textit{a} can be transliterated into Korean
graphemes on the basis of its context, like `ei' in the context of
\textit{-\underline{a}tion} and `a' in the context of
\textit{\underline{a}rt}.
    When context information is used,
determining the context window size is important.
    A context window that is too narrow can degrade transliteration performance because
of a lack of context information.
    For example, when English
grapheme \textit{t} in \textit{-\underline{t}ion} is transliterated
into Korean, the one right English grapheme is insufficient as
context because the three right contexts, \textit{-ion}, are
necessary to get the correct Korean grapheme, `s'.
    A context window that is too wide can also degrade
transliteration performance because it reduces the power to resolve
transliteration ambiguities.
    Many previous studies have determined that an appropriate context window size is 3.
    In this paper, we use a window size of 3, as in previous
work~\cite{kangbj00,goto03}.
    The effect of the context window size
on transliteration performance will be discussed in Section
\ref{sec:exp}.

\begin{table}[ht]
\begin{center}
\begin{tabular}{|l|l|c|c|c|c|c|c|c|c|c|}
 \hline
 & $Type$ &L3& L2& L1& C0& R1& R2& R3& &Output\\ \hline \hline
    \multirow{3}{*}{$\phi_{SP}$}  &$f_S$     &\$   &\$   &\$   &b   &o   &a
    &r & \multirow{3}{*}{$\rightarrow$}&\multirow{3}{*}{/B/}
    \\ \cline{3-9}
             &$f_{Stype}$  &\$   &\$   &\$   &C   &V   &V   &C   &
             &  \\ \cline{3-9}
             &$f_{P}$   &\$   &\$   &\$   & \multicolumn{4}{|c|}{$\epsilon$} & & \\ \cline{1-5} \hline \hline
    \multirow{3}{*}{$\phi_{ST}$}  &$f_S$     &\$   &\$   &\$   &b   &o   &a   &r   &
                    \multirow{3}{*}{$\rightarrow$}&\multirow{3}{*}{`b'} \\ \cline{3-9}
             &$f_{Stype}$  &\$   &\$   &\$   &C   &V   &V   &C   &   &
             \\ \cline{3-9}
             &$f_{T}$   &\$   &\$   &\$   &    \multicolumn{4}{|c|}{$\epsilon$}
                & &  \\ \hline \hline
    \multirow{3}{*}{$\phi_{PT}$} &$f_P$     &\$   &\$   &\$   &/B/ &/AO/ &/$\sim$/ &/R/ & \multirow{3}{*}{$\rightarrow$}&
                    \multirow{3}{*}{`b'}\\ \cline{3-9}
                &$f_{Ptype}$  &\$   &\$   &\$   &C   &V   &/$\sim$/ &C & &
                \\ \cline{3-9}
                &$f_{T}$   &\$   &\$   &\$   & \multicolumn{4}{|c|}{$\epsilon$}
                & &  \\ \hline \hline
    \multirow{5}{*}{$\phi_{(SP)T}$}    &$f_S$     &\$   &\$   &\$   &b   &o   &a   &r   &
            \multirow{5}{*}{$\rightarrow$}&\multirow{5}{*}{`b'} \\ \cline{3-9}
     &$f_P$     &\$   &\$   &\$   &/B/ &/AO/ &/$\sim$/ &/R/ &  &
     \\ \cline{3-9}
     &$f_{Stype}$  &\$   &\$   &\$   &C   &V   &V   &C   &    &
     \\ \cline{3-9}
     &$f_{Ptype}$  &\$   &\$   &\$   &C   &V   &/$\sim$/ &C &  &
     \\ \cline{3-9}
     &$f_{T}$   &\$   &\$   &\$   &  \multicolumn{4}{|c|}{$\epsilon$}
                & &  \\ \hline
\end{tabular}
\caption{\label{tab4}Framework for each component function: \$
represents start of words and $\epsilon$ means unused contexts for
each component function.}
\end{center}
\end{table}

    Table~\ref{tab4} shows how to identify the most relevant output in
each component function using context information.
    The L3-L1, C0, and R1-R3 represent the left context, current context
(i.e., that to be transliterated), and right context, respectively.
    The $\phi_{SP}$ function produces the most relevant source phoneme for
each source grapheme.
    If $SW=s_1\cdot s_2\cdot \ldots \cdot s_n$
is an English word, \textit{SW}'s pronunciation can be represented
as a sequence of source phonemes produced by $\phi_{SP}$; that is,
$P_{SW}=p_1\cdot p_2\cdot \ldots \cdot p_n$, where $p_i=
\phi_{SP}(s_i, c(s_i))$.
    $\phi_{SP}$ transforms source graphemes into phonemes in two ways.
    The first one is to search in a pronunciation dictionary containing
English words and their pronunciation~\cite{cmu97}.
    The second one is to estimate the pronunciation (or automatic
grapheme-to-phoneme
conversion)~\cite{andersen96,daeleman96,pagel98,damper99,chen03}.
    If an English word is not registered in the pronunciation
dictionary, we must estimate its pronunciation.
    The produced pronunciation is used for $\phi_{PT}$ in $\psi_{P}$ and
$\phi_{(SP)T}$ in $\psi_{C}$.
    For training the automatic grapheme-to-phoneme
conversion in $\phi_{SP}$, we use \textit{The CMU Pronouncing
Dictionary}~\cite{cmu97}.

    The $\phi_{ST}$, $\phi_{PT}$, and $\phi_{(SP)T}$ functions produce
target graphemes using their input.
    Like $\phi_{SP}$, these three functions use their previous outputs,
which are represented by $f_T$.
    As shown in Table~\ref{tab4},
$\phi_{ST}$, $\phi_{PT}$, and $\phi_{(SP)T}$ produce target grapheme
`b' for source grapheme \textit{b} and source phoneme /B/ in
\textit{\underline{b}oard} and /\underline{B} AO R D/.
    Because the \textit{b} and /B/ are the first source grapheme of
\textit{\underline{b}oard} and the first source phoneme of
/\underline{B} AO R D/, respectively, their left context is \$,
which represents the start of words.
    Source graphemes (\textit{o},
\textit{a}, and \textit{r}) and their type (V: vowel, V: vowel, and
C: consonant) can be the right context in $\phi_{ST}$ and
$\phi_{(SP)T}$.
    Source phonemes (/AO/, /$\sim$/, and /R/) and
their type (V: vowel, /$\sim$/: silence, V: vowel) can be the right
context in $\phi_{PT}$ and $\phi_{(SP)T}$.
    Depending on the feature type used in each component function and described in
Table~\ref{tab2}, $\phi_{ST}$, $\phi_{PT}$, and $\phi_{(SP)T}$
produce a sequence of target graphemes, $T_{SW}=t_1\cdot t_2\cdot
\ldots \cdot t_n$, for $SW=s_1\cdot s_2\cdot \ldots \cdot s_n$ and
$P_{SW}=p_1\cdot p_2\cdot \ldots \cdot p_n$.
    For \textit{board}, $SW$, $P_{SW}$, and $T_{SW}$ can be represented as follows.
    The /$\sim$/ represents silence (null source phonemes), and the
`$\sim$' represents null target graphemes.
\begin{itemize}
    \item $SW= s_1\cdot s_2\cdot s_3\cdot s_4\cdot s_5 = b \cdot o \cdot a \cdot r \cdot d$
    \item $P_{SW}= p_1\cdot p_2\cdot p_3\cdot p_4\cdot p_5 = /B/\cdot /AO/ \cdot /\sim/ \cdot /R/ \cdot /D/$
    \item $T_{SW}= t_1\cdot t_2\cdot t_3\cdot t_4\cdot t_5 =$ `b'$\cdot$ `o' $\cdot$ `$\sim$' $\cdot$ `$\sim$' $\cdot$ `deu'
\end{itemize}

\subsection{Machine Learning Algorithms for Each Component
Function} \label{subsec:ml}

    In this section we describe a way to model component functions
using three machine learning algorithms (the maximum entropy model,
decision-tree learning, and memory-based learning)\footnote{These
three algorithms are typically applied to automatic
grapheme-to-phoneme
conversion~\cite{andersen96,daeleman96,pagel98,damper99,chen03}.}.
    Because the four component functions share a similar framework, we
limit our focus to $\phi_{(SP)T}$ in this section.

\subsubsection{Maximum entropy model}
    The maximum entropy model (MEM) is a widely used probability model
that can incorporate heterogeneous information
effectively~\cite{berger96}.
    In the MEM, an event ($ev$) is usually composed of a target event ($te$)
and a history event ($he$); say $ev=<te, he>$.
    Event $ev$ is represented by a bundle of feature functions, $fe_i(ev)$, which
represent the existence of certain characteristics in event $ev$.
    A feature function is a binary-valued function.
    It is activated ($fe_i(ev)=1$) when it meets its activating condition; otherwise
it is deactivated ($fe_i(ev)=0$)~\cite{berger96}.
    Let source language word \textit{SW} be composed of \textit{n} graphemes.
\textit{SW}, $P_{SW}$, and $T_{SW}$ can then be represented as
$SW=s_1,\cdots,s_n$, $P_{SW}=p_1,\cdots,p_n$, and
$T_{SW}=t_1,\cdots,t_n$, respectively.
    $P_{SW}$ and $T_{SW}$ represent the pronunciation and
target language word corresponding to \textit{SW}, and $p_i$ and
$t_i$ represent the source phoneme and target grapheme corresponding
to $s_i$.
    Function $\phi_{(SP)T}$ based on the maximum entropy model can be represented as
\begin{eqnarray}
\label{eqn4:0}
 Pr(T_{SW}|SW,P_{SW})
 = Pr(t_1,\cdots,t_n|s_1,\cdots,s_n, p_1,\cdots,p_n)
\end{eqnarray}
    With the assumption that $\phi_{(SP)T}$ depends on the context
information in window size \textit{k}, we simplify
Formula~(\ref{eqn4:0}) to
\begin{eqnarray}
\label{eqn4:1}
 Pr(T_{SW}|SW,P_{SW})
 \approx \prod_{i} Pr(t_i|t_{i-k},\cdots, t_{i-1}, p_{i-k},\cdots, p_{i+k}, s_{i-k},\cdots, s_{i+k})
\end{eqnarray}
    Because $t_1,\cdots,t_n$, $s_1,\cdots,s_n$, and $p_1,\cdots,p_n$
can be represented by $f_T$, $f_{S,Stype}$, and $f_{P,Ptype}$,
respectively, we can rewrite Formula~(\ref{eqn4:1}) as
\begin{eqnarray}
\label{eqn4:2}
 Pr(T_{SW}|SW,P_{SW})
 \approx \prod_{i} Pr(t_i|f_{T_{(i-k,i-1)}}, f_{P,Ptype_{(i-k,i+k)}}, f_{S,Stype_{(i-k,i+k)}})
\end{eqnarray}
    where $i$ is the index of the current source grapheme and source
phoneme to be transliterated and $f_{X(l,m)}$ represents the
features of feature type $f_X$ located from position $l$ to position
$m$.

    An important factor in designing a model based on the maximum
entropy model is to identify feature functions that effectively
support certain decisions of the model.
    Our basic philosophy of
feature function design for each component function is that the
context information collocated with the unit of interest is
important.
    We thus designed the feature function with collocated
features in each feature type and between different feature types.
    Features used for $\phi_{(SP)T}$ are listed below.
    These features are used as activating conditions or history events of feature
functions.
\begin{itemize}
 \item[$\bullet$] Feature type and features used for designing feature functions in $\phi_{(SP)T}$ ($k=3$)
 \begin{itemize}
 \item All possible features in $f_{S,Stype_{i-k,i+k}}$, $f_{P,Ptype_{i-k,i+k}}$, and $f_{T_{i-k,i-1}}$ (e.g., $f_{S_{i-1}}$, $f_{P_{i-1}}$, and $f_{T_{i-1}}$)
 \item All possible feature combinations between features of the same feature type (e.g., \{$f_{S_{i-2}}$, $f_{S_{i-1}}$, $f_{S_{i+1}}$\}, \{$f_{P_{i-2}}$, $f_{P_{i}}$, $f_{P_{i+2}}$\}, and \{$f_{T_{i-2}}$, $f_{T_{i-1}}$\})
 \item All possible feature combinations between features of different feature types (e.g., \{$f_{S_{i-1}}$, $f_{P_{i-1}}$\}, \{$f_{S_{i-1}}$, $f_{T_{i-2}}$\} , and \{$f_{Ptype_{i-2}}$, $f_{P_{i-3}}$, $f_{T_{i-2}}$\})
 \begin{itemize}
 \item between $f_{S,Stype_{i-k,i+k}}$ and $f_{P,Ptype_{i-k,i+k}}$
 \item between $f_{S,Stype_{i-k,i+k}}$ and $f_{T_{i-k,i-1}}$
 \item between $f_{P,Ptype_{i-k,i+k}}$ and $f_{T_{i-k,i-1}}$
 \end{itemize}
 \end{itemize}
\end{itemize}

    Generally, a conditional maximum entropy model that gives the
conditional probability $Pr(y|x)$ is represented as
Formula~(\ref{eqn4:3-0})~\cite{berger96}.
\begin{eqnarray}
\label{eqn4:3-0}
 Pr(y|x) &=& \frac{1}{Z(x)} exp(\sum_i \lambda_i fe_i(x,y)) \\ \nonumber
  Z(x) &=& \sum_{y} exp(\sum_i \lambda_i fe_i(x,y))
\end{eqnarray}
    In $\phi_{(SP)T}$, the target event ($te$) is target graphemes to
be assigned, and the history event ($he$) can be represented as a
tuple $<f_{T_{(i-k,i-1)}},$ $f_{S,Stype_{(i-k,i+k)}},$
$f_{P,Ptype_{(i-k,i+k)}}>$.
    Therefore, we can rewrite Formula~(\ref{eqn4:2}) as
\begin{eqnarray}
\label{eqn4:3}
 &&Pr(t_i|f_{T_{(i-k,i-1)}}, f_{S,Stype_{(i-k,i+k)}},
 f_{P,Ptype_{(i-k,i+k)}}) \\ \nonumber
 &&= Pr (te|he) = \frac{1}{Z(he)} exp(\sum_i \lambda_i fe_i(he,te))
\end{eqnarray}

\begin{table}[tb]
\medskip
\begin{center}

\begin{tabular}{|c|c|c|c|c|}
\hline
 \multirow{2}{*}{$fe_j$} & $te$ & \multicolumn{3}{c|}{$he$} \\
 \cline{2-5}
                         & $t_i$ & $f_{T_{(i-k,i-1)}}$ &
 $f_{S,Stype_{(i-k,i+k)}}$ & $f_{P,Ptype_{(i-k,i+k)}}$ \\ \hline
 \hline
 $fe_1$ & `b' & -- & $f_{S_i} = b$ & $f_{P_i}=/B/$ \\
 $fe_2$ & `b' & -- & $f_{S_{i-1}} = \$$ & -- \\
 $fe_3$ & `b' & $f_{T_{i-1}} = \$ $ & $f_{S_{i+1}} = o$ and $f_{Stype_{i+2}} = V$ & $f_{P_i}=/B/$ \\
 $fe_4$ & `b' & -- & -- & $f_{P_{i+1}}=/AO/$ \\
 $fe_5$ & `b' & $f_{T_{i-2}} = \$ $ & $f_{S_{i+3}} = r$ & $f_{Ptype_i}=C$
 \\ \hline
 \end{tabular}
\caption{\label{tab4-1} Feature functions for $\phi_{(SP)T}$ derived
from Table~\ref{tab4}.}
\end{center}
\end{table}
    Table~\ref{tab4-1} shows example feature functions for
$\phi_{(SP)T}$; Table~\ref{tab4} was used to derive the functions.
    For example, $fe_1$ represents an event where $he$ (history event)
is ``$f_{S_i}$ is \textit{b} and $f_{P_i}$ is /B/" and $te$ (target
event) is ``$f_{T_i}$ is `b'".
    To model each component
function based on the MEM, Zhang's maximum entropy modeling tool is
used~\cite{zhang04}.

\subsubsection{Decision-tree learning}

    Decision-tree learning (DTL) is one of the most widely used and
well-known methods for inductive
inference~\cite{quinlan86,mitchell97}.
    ID3, which is a greedy
algorithm that constructs decision trees in a top-down manner, uses
the information gain, which is a measure of how well a given feature
(or attribute) separates training examples on the basis of their
target class~\cite{quinlan93,manning99}.
    We use C4.5~\cite{quinlan93}, which is a well-known tool for
DTL and an implementation of Quinlan's ID3 algorithm.

\begin{figure}[ht]
\begin{center}
\includegraphics[width=0.600\textwidth]{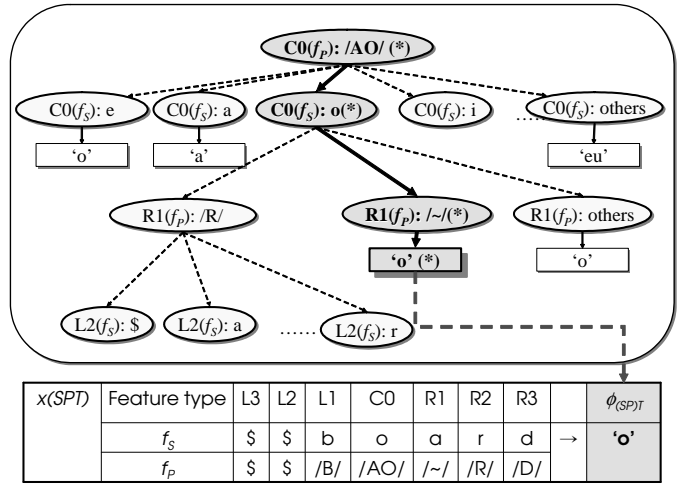}
\end{center}
\caption{\label{fig4} Decision tree for $\phi_{(SP)T}$.}
\end{figure}

    The training data for each component function is represented by
features located in L3-L1, C0, and R1-R3, as shown in
Table~\ref{tab4}.
    C4.5 tries to construct a decision tree by
looking for regularities in the training data~\cite{mitchell97}.
    Figure~\ref{fig4} shows part of the decision tree constructed for
$\phi_{(SP)T}$ in English-to-Korean transliteration.
    A set of the target classes in the decision tree for $\phi_{(SP)T}$ is a set of
the target graphemes.
    The rectangles indicate the leaf nodes,
which represent the target classes, and the circles indicate the
decision nodes.
    To simplify our examples, we use only $f_S$ and
$f_P$.
    Note that all feature types for each component function, as
described in Table~\ref{tab4}, are actually used to construct
decision trees.
    Intuitively, the most effective feature from
among L3-L1, C0, and R1-R3 for $\phi_{(SP)T}$ may be located in C0
because the correct outputs of $\phi_{(SP)T}$ strongly depend on the
source grapheme or source phoneme in the C0 position.
    As we expected, the most effective feature in the decision tree is
located in the C0 position, that is, C0($f_P$).
    (Note that the
first feature to be tested in decision trees is the most effective
feature.)
    In Figure~\ref{fig4}, the decision tree produces the
target grapheme (Korean grapheme) `o' for the instance $x(SPT)$ by
retrieving the decision nodes from $C0(f_P)=/AO/$ to
$R1(f_P)=/\sim/$ represented by `$\ast$'.

\subsubsection{Memory-based learning}

    Memory-based learning (MBL), also called ``instance-based learning"
and ``case-based learning", is an example-based learning method.
    It is based on a $k$-nearest neighborhood
algorithm~\cite{Aha91,Aha97,Cover67,Devijver82}.
    MBL represents training data as a vector and, in the training phase, it places
all training data as examples in memory and clusters some examples
on the basis of the $k$-nearest neighborhood principle.
    Training data for MBL is represented in the same form as training data for
a decision tree.
    Note that the target classes for $\phi_{(SP)T}$,
which MBL outputs, are target graphemes.
    Feature weighting to deal
with features of differing importance is also done in the training
phase\footnote{TiMBL~\cite{daelemans04} supports \textit{gain ratio
weighting, information gain weighting, chi-squared ($\chi^2$)
weighting}, and \textit{shared variance weighting} of the
features.}.
    It then produces an output using similarity-based
reasoning between test data and the examples in memory.
    If the test data is $x$ and the set of examples in memory is $Y$, the
similarity between $x$ and $Y$ can be estimated using distance
function $\Delta(x,Y)$\footnote{\textit{Modified value difference
metric, overlap metric, Jeffrey divergence metric}, \textit{dot
product metric}, etc. are used as the distance
function~\cite{daelemans04}.}.
    MBL selects an example $y_i$ or the
cluster of examples that are most similar to $x$ and then assigns
the example's target class to $x$'s target class.
    We use an MBL tool called TiMBL (Tilburg memory-based learner) version
5.0~\cite{daelemans04}.


\section{Experiments}
\label{sec:exp}

    We tested the four machine transliteration models
on English-to-Korean and English-to-Japanese transliteration.
    The test set for the former (EKSet)~\cite{nam97} consisted of 7,172
English-Korean pairs -- the number of training items was about 6,000
and that of the blind test items was about 1,000.
    EKSet contained no transliteration variations, meaning that there was
one transliteration for each English word.
    The test set for the latter (EJSet) contained English-katakana pairs from
EDICT~\cite{breen03} and consisted of 10,417 pairs -- the number of
training items was about 9,000 and that of the blind test items was
about 1,000.
    EJSet contained transliteration variations, like
$<$\textit{micro}, `ma-i-ku-ro'$>$, and $<$\textit{micro},
`mi-ku-ro'$>$; the average number of Japanese transliterations for
an English word was 1.15.
    EKSet and EJSet covered proper names, technical terms,
and general terms.
    We used \textit{The CMU Pronouncing
Dictionary}~\cite{cmu97} for training pronunciation estimation (or
automatic grapheme-to-phoneme conversion) in $\phi_{SP}$.
    The training for automatic grapheme-to-phoneme conversion
was done ignoring the lexical stress of vowels in the
dictionary~\cite{cmu97}.
    The evaluation was done in terms of word
accuracy ($WA$), the evaluation measure used in previous
work~\cite{kangbj00,kangih00,goto03,bilac04}.
    Here, $WA$ can be represented as Formula~(\ref{eqn4}).
    A generated transliteration for an English word was judged to be correct
if it exactly matched a transliteration for that word in the test
data.
\begin{eqnarray}
\label{eqn4} WA &=& \frac{number\ of\ correct\ transliterations\
output\ by\ system}{number \ of\ transliterations\ in\ blind\ test\
data}
\end{eqnarray}
    In the evaluation, we used $k$-fold cross-validation ($k$=7 for
EKSet and $k$=10 for EJSet).
    The test set was divided into $k$ subsets.
    Each was used in turn for testing while the remainder was
used for training.
    The average $WA$ computed across all $k$ trials
was used as the evaluation results presented in this section.

We conducted six tests.
\begin{itemize}
 \item \textsf{Hybrid Model Test}:
    Evaluation of hybrid transliteration model by changing value
    of $\alpha$ (the parameter of the hybrid transliteration model)
 \item \textsf{Comparison Test I}:
    Comparison among four machine transliteration models
 \item \textsf{Comparison Test II}:
    Comparison of four machine transliteration models to previously proposed transliteration methods
 \item \textsf{Dictionary Test}:
    Evaluation of transliteration models on words registered and not registered
    in pronunciation dictionary to determine effect of pronunciation dictionary
    on each model
 \item \textsf{Context Window-Size Test}:
    Evaluation of transliteration models for various sizes of context window
 \item \textsf{Training Data-Size Test}:
    Evaluation of transliteration models for various sizes of training data sets
\end{itemize}

\subsection{Hybrid Model Test}
\label{sect:exp_hybrid}

    The objective of this test was to estimate the dependence of the
performance of $\psi_{H}$ on parameter $\alpha$.
    We evaluated the performance by changing $\alpha$ from 0 to 1 at intervals of 0.1
(i.e., $\alpha$=0, 0.1, 0.2, $\cdots$, 0.9, 1.0).
    Note that the
hybrid model can be represented as ``$\alpha \times Pr(\psi_{P}) +
(1-\alpha) \times Pr(\psi_{G})$".
    Therefore, $\psi_{H}$ is $\psi_{G}$ when $\alpha=0$ and $\psi_{P}$ when $\alpha=1$.
    As shown in Table~\ref{tab4-2}, the performance of $\psi_{H}$
depended on that of $\psi_{G}$ and $\psi_{P}$.
    For example, the performance of $\psi_{G}$ exceeded that of $\psi_{P}$ for EKSet.
    Therefore, $\psi_{H}$ tended to perform better when $\alpha \leq
0.5$ than when $\alpha > 0.5$ for EKSet.
    The best performance was attained when $\alpha=0.4$ for EKSet and
when $\alpha=0.5$ for EJSet.
    Hereinafter, we use $\alpha=0.4$ for
EKSet and $\alpha=0.5$ for EJSet as the linear interpolation
parameter for $\psi_{H}$.

\begin{table}[ht]
\begin{center}
\begin{tabular}{|l|c|c|}
\hline
    $\alpha$     & EKSet &  EJSet  \\ \hline \hline
    0           & 58.8\% & 58.8\%  \\ \hline
     0.1           & 61.2\% & 60.9\%  \\ \hline
     0.2           & 62.0\% & 62.6\%  \\ \hline
     0.3           & 63.0\% & 64.1\%  \\ \hline
     0.4           & 64.1\% & 65.4\%  \\ \hline
     0.5           & 63.4\% & 65.8\%  \\ \hline
     0.6           & 61.1\% & 65.0\%  \\ \hline
     0.7           & 59.6\% & 63.4\%  \\ \hline
     0.8           & 58.2\% & 62.1\%  \\ \hline
     0.9           & 57.0\% & 61.2\%  \\ \hline
     1.0           & 55.2\% & 59.2\%  \\ \hline
\end{tabular}
\caption{\label{tab4-2} Results of Hybrid Model Test.}
\end{center}
\end{table}

\subsection{Comparison Test I} \label{sect:exp_comp1}

    The objectives of the first comparison test were to compare
performance among the four transliteration models ($\psi_{G}$,
$\psi_{P}$, $\psi_{H}$, and $\psi_{C}$) and to compare the
performance of each model with the combined performance of three of
the models ($\psi_{G+P+C}$).
    Table~\ref{tab5} summarizes the
performance of each model for English-to-Korean and
English-to-Japanese transliteration, where DTL, MBL\footnote{We
tested all possible combinations between $\Delta(x,Y)$ and a
weighting scheme supported by TiMBL~\cite{daelemans04} and did not
detect any significant differences in performance for the various
combinations. Therefore, we used the default setting of TiMBL
(\textit{Overlap metric} for $\Delta(x,Y)$ and \textit{gain ratio
weighting} for feature weighting).} and MEM represent decision-tree
learning, memory-based learning, and maximum entropy model.

    The unit to be transliterated was restricted to either a source
grapheme or a source phoneme in $\psi_{G}$ and $\psi_{P}$; it was
dynamically selected on the basis of the contexts in $\psi_{H}$ and
$\psi_{C}$.
    This means that $\psi_{G}$ and $\psi_{P}$ could
produce an incorrect result if either a source phoneme or a source
grapheme, which, respectively, they do not consider, holds the key
to producing the correct transliteration result.
    For this reason, $\psi_{H}$ and $\psi_{C}$ performed better than both $\psi_{G}$
and $\psi_{P}$.

\begin{table}[ht]
\begin{center}
\begin{tabular}{|l|c|c|c|c|c|c|}
\hline
\multirow{2}{*}{Transliteration Model} &\multicolumn{3}{|c|}{EKSet} & \multicolumn{3}{|c|}{EJSet} \\
  \cline{2-7}

        &DTL &MBL  &MEM &DTL &MBL  &MEM \\ \hline \hline
    $\psi_{G}$   &53.1\% &54.6\%  &58.8\%   &55.6\%   &58.9\% &58.8\% \\ \hline
    $\psi_{P}$   &50.8\%   &50.6\%  &55.2\% &55.8\%   &56.1\% &59.2\% \\ \hline
    $\psi_{H}$      &N/A &N/A  &64.1\% & N/A &N/A & 65.8\% \\ \hline
    $\psi_{C}$      &59.5\% &60.3\%  &65.5\% &64.0\%   &65.8\%   &69.1\% \\ \hline
    $\psi_{G+P+C}$ & 72.0\%   &71.4\%  &75.2\% &73.4\%   &74.2\%   &76.6\% \\
\hline
\end{tabular}
\caption{\label{tab5} Results of Comparison Test I.}
\end{center}
\end{table}

    In the table, $\psi_{G+P+C}$ means the combined results for the
three transliteration models, $\psi_{G}$, $\psi_{P}$, and
$\psi_{C}$.
    We exclude $\psi_{H}$ from the combining because it is
implemented only with the MEM (the performance of combining the four
transliteration models are discussed in Section~\ref{sec:discuss}).
    In evaluating $\psi_{G+P+C}$, we
judged the transliteration results to be correct if there was at
least one correct transliteration among the results produced by the
three models.
    Though $\psi_{C}$ showed the best results among
the three transliteration models due to its ability to use the
correspondence between the source grapheme and source phoneme, the
source grapheme or the source phoneme can create noise when the
correct transliteration is produced by the other one.
    In other words, when the correct transliteration is strongly
biased to either grapheme-based or phoneme-based transliteration,
$\psi_{G}$ and $\psi_{P}$ may be more suitable for producing the
correct transliteration.

    Table~\ref{tab6} shows example transliterations produced by each
transliteration model.
    The $\psi_{G}$ produced correct
transliterations for \textit{cyclase} and \textit{bacteroid}, while
$\psi_{P}$ did the same for \textit{geoid} and \textit{silo}.
    $\psi_{C}$ produced correct transliterations for
\textit{saxhorn} and \textit{bacteroid}, and $\psi_{H}$ produced
correct transliterations for \textit{geoid} and \textit{bacteroid}.
    As shown by these results, there are
transliterations that only one transliteration model can produce
correctly.
    For example, only $\psi_{G}$, $\psi_{P}$, and
$\psi_{C}$ produced the correct transliterations of
\textit{cyclase}, \textit{silo}, and \textit{saxhorn}, respectively.
    Therefore, these three transliteration models
can be used in a complementary manner to improve transliteration
performance because at least one can usually produce the correct
transliteration.
    This combination increased the performance by compared to
$\psi_{G}$, $\psi_{P}$, and $\psi_{C}$ (on average, 30.1\% in EKSet
and 24.6\% in EJSet).
    In short, $\psi_{G}$,
$\psi_{P}$, and $\psi_{C}$ are complementary transliteration models
that together produce more correct transliterations, so combining
different transliteration models can improve transliteration
performance.
    The transliteration results produced
by $\psi_{G+P+C}$ are analyzed in detail in
Section~\ref{sec:discuss}.

\begin{table}[ht]
\begin{center}
\begin{tabular}{|c|c|c|}
 \hline
         & $\psi_{G}$ &$\psi_{P}$ \\ \hline \hline
    \textit{cyclase} &`\textsf{si-keul-la-a-je}'
    &$\ast$`sa-i-keul-la-a-je' \\ \hline
    \textit{bacteroid} & `\textsf{bak-te-lo-i-deu}'
    &$\ast$`bak-teo-o-i-deu' \\ \hline
    \textit{geoid}  & $\ast$`je-o-i-deu' & `\textsf{ji-o-i-deu}' \\
\hline
    \textit{silo}  &$\ast$`sil-lo' &`\textsf{sa-il-lo}' \\
\hline
\textit{saxhorn}&$\ast$`saek-seon' &$\ast$`saek-seu-ho-leun' \\
    \hline \hline
     & $\psi_{H}$ &$\psi_{C}$   \\ \hline  \hline
    \textit{cyclase}   &$\ast$`sa-i-keul-la-a-je' &$\ast$`sa-i-keul-la-a-je'  \\ \hline
    \textit{bacteroid}    &`\textsf{bak-te-lo-i-deu}'
    &`\textsf{bak-te-lo-i-deu}'  \\ \hline
    \textit{geoid} & `\textsf{ji-o-i-deu}'& $\ast$`ge-o-i-deu' \\
    \hline
    \textit{silo}    &$\ast$`sil-lo' & $\ast$`sil-lo'  \\
    \hline
    \textit{saxhorn}    &$\ast$`saek-seon' &`\textsf{saek-seu-hon}'  \\
\hline
\end{tabular}
\caption{\label{tab6} Example transliterations produced by each
transliteration model ($\ast$ indicates an incorrect
transliteration).}
\end{center}
\end{table}

    In our subsequent testing, we used the maximum entropy model as
the machine learning algorithm for two reasons.
    First, it produced the best results of the three algorithms we tested\footnote{A
one-tail paired t-test showed that the results with the MEM were
always significantly better (except for $\phi_G$ in EJSet) than
those of DTL and MBL (level of significance = 0.001).}.
    Second, it can support $\psi_H$.

\subsection{Comparison Test II}
\label{sect:exp_comp2}

    In this test, we compared four previously proposed machine
transliteration methods~\cite{kangbj00,kangih00,goto03,bilac04} to
the four transliteration models ($\psi_G$, $\psi_P$, $\psi_H$, and
$\psi_C$), which were based on the MEM.
    Table~\ref{tab7} shows the results.
    We trained and tested the previous methods with the same data
sets used for the four transliteration models.
    Table~\ref{tab8} shows the key features of the methods and models from the
viewpoint of information type and usage.
    Information type indicates the type
of information considered: source grapheme, source phoneme, and
correspondence between the two.
    For example, the first three methods use only the source grapheme.
    Information usage indicates the context used and whether the previous output
is used.

\begin{table}[ht]
\begin{center}
\begin{tabular}{|c|c|c|c|}
  \hline
\multicolumn{2}{|c|}{Method/Model} &  {EKSet} &{EJSet} \\ \hline
\hline
\multirow{4}{*}{Previous methods} & Kang and Choi~\citeyear{kangbj00}  &  51.4\%&     50.3\% \\
    \cline{2-4}
    & Kang and Kim~\citeyear{kangih00}    &55.1\%&     53.2\%    \\ \cline{2-4}
    & Goto \textit{et al.}~\citeyear{goto03}  &55.9\%&     56.2\%    \\ \cline{2-4}
    & Bilac and Tanaka~\citeyear{bilac04}   &58.3\%&     62.5\%    \\ \hline
\multirow{4}{*}{MEM-based models} & $\psi_{G}$ &58.8\%&     58.8\% \\
\cline{2-4}
    &$\psi_{P}$ &55.2\%&      59.2\% \\ \cline{2-4}
    &$\psi_{H}$ &64.1\%&      65.8\% \\ \cline{2-4}
    &$\psi_{C}$ &65.5\%&      69.1\% \\ \hline
\end{tabular}
\caption{\label{tab7}Results of Comparison Test II.}
\end{center}
\end{table}

\begin{table}[ht]
\begin{center}
\begin{tabular}{|c|c|c|c|c|c|}
  \hline
\multirow{2}{*}{Method/Model}  &  \multicolumn{3}{c|}{Info. type}
&\multicolumn{2}{c|}{Info. usage}
\\ \cline{2-6}
      & S   & P   & C   &Context    & PO \\ \hline \hline
 Kang and Choi~\citeyear{kangbj00}     &+   &--   &--   &$<-3\sim+3>$    &-- \\ \hline
 Kang and Kim~\citeyear{kangih00}    &+   &--   &--   &Unbounded     &+ \\ \hline
 Goto \textit{et al.}~\citeyear{goto03}   &+   &--   &--   &$<-3\sim+3>$    &+ \\ \hline
 Bilac and Tanaka~\citeyear{bilac04}  &+   &+   &--   & Unbounded             &-- \\ \hline
 $\psi_{G}$    &+   &--   &--   &$<-3\sim+3>$    &+ \\ \hline
 $\psi_{P}$    &--   &+   &--   &$<-3\sim+3>$    &+ \\ \hline
 $\psi_{H}$    &+   &+   &--   &$<-3\sim+3>$    &+ \\ \hline
 $\psi_{C}$    &+   &+   &+   &$<-3\sim+3>$    &+ \\ \hline
\end{tabular}
\caption{\label{tab8} Information type and usage for previous
methods and four transliteration models, where S, P, C, and PO
respectively represent the source grapheme, source phoneme,
correspondence between S and P, and previous output.}
\end{center}
\end{table}

    It is obvious from the table that the more information types a
transliteration model considers, the better its performance.
    Either the source phoneme or the correspondence -- which are not
considered in the methods of Kang and Choi~\citeyear{kangbj00}, Kang
and Kim~\citeyear{kangih00}, and Goto~\textit{et
al.}~\citeyear{goto03} -- is the key to the higher performance of
the method of Bilac and Tanaka~\citeyear{bilac04} and the $\psi_{H}$
and $\psi_{C}$.

    From the viewpoint of information usage, the models and methods
that consider the previous output tended to achieve better
performance.
    For example, the method of Goto~\textit{et
al.}~\citeyear{goto03} had better results than that of Kang and
Choi~\citeyear{kangbj00}.
    Because machine transliteration is
sensitive to context, a reasonable context size usually enhances
transliteration ability.
    Note that the size of the context window
for the previous methods was limited to 3 because a context window
wider than 3 degrades performance ~\cite{kangbj00} or does not
significantly improve it~\cite{kangih00}.
    Experimental results related to context window size are given in
Section~\ref{sec:exp_context}.

    Overall, $\psi_{H}$ and $\psi_{C}$ had better performance than the
previous methods (on average, 17.04\% better for EKSet and 21.78\%
better for EJSet), $\psi_{G}$ (on average, 9.6\% better for EKSet
and 14.4\% better for EJSet), and $\psi_{P}$ (on average, 16.7\%
better for EKSet and 19.0\% better for EJSet).
    In short, a good machine transliteration model should 1) consider either the
correspondence between the source grapheme and the source phoneme or
both the source grapheme and the source phoneme, 2) have a
reasonable context size, and 3) consider previous output.
    The $\psi_{H}$ and $\psi_{C}$ satisfy all three conditions.

\subsection{Dictionary Test}
\label{sect:exp_dic}

    Table~\ref{tab9} shows the performance of each transliteration
model for the dictionary test.
    In this test, we evaluated four transliteration models according
to a way of pronunciation generation (or grapheme-to-phoneme
conversion).
    \textit{Registered} represents the performance for words
registered in the pronunciation dictionary, and
\textit{Unregistered} represents that for unregistered words.
    On average, the number of \textit{Registered} words in EKSet was about 600,
and that in EJSet was about 700 in $k$-fold cross-validation test
data.
    In other words, \textit{Registered} words accounted for about 60\% of the test
data in EKSet and about 70\% of the test data in EJSet.
    The correct pronunciation can always be acquired from the
pronunciation dictionary for \textit{Registered} words, while the
pronunciation must be estimated for \textit{Unregistered} words
through automatic grapheme-to-phoneme conversion.
    However, the automatic grapheme-to-phoneme conversion does not always produce correct
pronunciations -- the estimated rate of correct pronunciations was
about 70\% accuracy.
\begin{table}[ht]
\begin{center}
\begin{tabular}{|c|c|c|c|c|}
  \hline
      &\multicolumn{2}{c|}{EKSet} &  \multicolumn{2}{c|}{EJSet}
    \\ \cline{2-5}
      &\textit{Registered}   &\textit{Unregistered}  &\textit{Registered}   &\textit{Unregistered}  \\
      \hline \hline
    $\psi_G$ &60.91\%&  55.74\%&  61.18\%&  50.24\%  \\ \hline
    $\psi_P$  &66.70\%&  38.45\%&  64.35\%&  40.78\%    \\ \hline
    $\psi_H$  &70.34\%&  53.31\%&  70.20\%&  50.02\% \\ \hline
    $\psi_C$ &73.32\%&  54.12\%&  74.04\%&  51.39\%  \\ \hline
    $ALL$ &80.78\%&  68.41\%&  81.17\%&  62.31\%  \\
    \hline
\end{tabular}
\caption{\label{tab9}Results of Dictionary Test: ALL means
$\psi_{G+P+H+C}$. }
\end{center}
\end{table}

    Analysis of the results showed that the four transliteration
models fall into three categories.
    Since the $\psi_G$ is free from the need for correct pronunciation,
that is, it does not use the source phoneme, its performance is not
affected by pronunciation correctness.
    Therefore, $\psi_G$ can be regarded as the baseline performance
for \textit{Registered} and \textit{Unregistered}.
    Because $\psi_{P}$ ($\phi_{PT} \circ \phi_{SP}$), $\psi_{H}$ ($\alpha
\times$ $Pr(\psi_P)$+$(1-\alpha)\times$ $Pr(\psi_G)$), and
$\psi_{C}$ ($\phi_{(SP)T} \circ \phi_{SP}$) depend on the source
phoneme, their performance tends to be affected by the performance
of $\phi_{SP}$.
    Therefore, $\psi_{P}$, $\psi_{H}$, and $\psi_{C}$
show notable differences in performance between \textit{Registered}
and \textit{Unregistered}.
    However, the performance gap differs with the strength of the dependence.
$\psi_P$ falls into the second category: its performance strongly
depends on the correct pronunciation.
    $\psi_P$ tends to perform well for \textit{Registered} and
poorly for \textit{Unregistered}.
    $\psi_H$ and $\psi_C$ weakly depend on the correct pronunciation.
    Unlike $\psi_P$, they make
use of both the source grapheme and source phoneme.
    Therefore, they can perform reasonably well without the correct pronunciation
because using the source grapheme weakens the negative effect of
incorrect pronunciation in machine transliteration.

    Comparing $\psi_{C}$ and $\psi_{P}$, we find two interesting things.
    First, $\psi_{P}$ was more sensitive to errors in
$\phi_{SP}$ for \textit{Unregistered}.
    Second, $\psi_{C}$ showed better results for both \textit{Registered} and
\textit{Unregistered}.
    Because $\psi_{P}$ and $\psi_{C}$ share the same function, $\phi_{SP}$,
the key factor accounting for the performance gap between them is
the component functions, $\phi_{PT}$ and $\phi_{(SP)T}$.
    From the results shown in Table~\ref{tab9}, we can infer that
$\phi_{(SP)T}$ (in $\psi_{C}$) performed better than $\phi_{PT}$ (in
$\psi_{P}$) for both \textit{Registered} and \textit{Unregistered}.
    In $\phi_{(SP)T}$, the source grapheme
corresponding to the source phonemes, which $\phi_{PT}$ does not
consider, made two contributions to the higher performance of
$\phi_{(SP)T}$.
    First, the source grapheme in the correspondence made it possible
to produce more accurate transliterations.
    Because $\phi_{(SP)T}$
considers the correspondence, $\phi_{(SP)T}$ has a more powerful
transliteration ability than $\phi_{PT}$, which uses just the source
phonemes, when the correspondence is needed to produce correct
transliterations.
    This is the main reason $\phi_{(SP)T}$
performed better than $\phi_{PT}$ for \textit{Registered}.
    Second, source graphemes in the correspondence compensated for errors
produced by $\phi_{SP}$ in producing target graphemes.
    This is the main reason $\phi_{(SP)T}$ performed better than $\phi_{PT}$ for
\textit{Unregistered}.
    In the comparison between $\psi_{C}$ and $\psi_{G}$,
the performances were similar for \textit{Unregistered}.
    This indicates that the transliteration power of $\psi_{C}$ is similar
to that of $\psi_{G}$, even though the pronunciation of the source
language word may not be correct.
    Furthermore, the performance of $\psi_{C}$ was significantly higher than
that of $\psi_{G}$ for \textit{Registered}.
    This indicates that
the transliteration power of $\psi_{C}$ is greater than that of
$\psi_{G}$ if the correct pronunciation is given.

    The behavior of $\psi_{H}$ was similar to that of $\psi_{C}$.
    For \textit{Unregistered}, $Pr(\psi_G)$ in $\psi_{H}$ made it possible
for $\psi_{H}$ to avoid errors caused by $Pr(\psi_P)$.
    Therefore, it worked better than $\psi_P$.
    For \textit{Registered}, $Pr(\psi_P)$ enabled $\psi_{H}$ to perform better than
$\psi_{G}$.

    The results of this test showed that $\psi_{H}$ and $\psi_{C}$
perform better than $\psi_{G}$ and $\psi_{P}$ while complementing
$\psi_{G}$ and $\psi_{P}$ (and thus overcoming their disadvantage)
by considering either the correspondence between the source grapheme
and the source phoneme or both the source grapheme and the source
phoneme.

\subsection{Context Window-Size Test}
\label{sec:exp_context}

    In our testing of the effect of the context window size, we varied
the size from 1 to 5.
    Regardless of the size, $\psi_{H}$ and $\psi_{C}$ always
performed better than both $\psi_{G}$ and $\psi_{P}$.
    When the size was 4 or 5, each model had difficulty identifying
regularities in the training data.
    Thus, there were consistent drops in performance for all models
when the size was increased from 3 to 4 or 5.
    Although the best performance was obtained when the size was 3, as shown in
Table~\ref{tab10}, the differences in performance were not
significant in the range of 2-4.
    However, there was a significant
difference between a size of 1 and a size of 2.
    This indicates that a lack of contextual information can easily lead to incorrect
transliteration.
    For example, to produce the correct target
language grapheme of \textit{t} in \textit{-tion}, we need the right
three graphemes (or at least the right two) of \textit{t},
\textit{-ion} (or \textit{-io}).
    The results of this testing
indicate that the context size should be more than 1 to avoid
degraded performance.

\begin{table}[ht]
\begin{center}
\begin{tabular}{|c|c|c|c|c|c|}
  \hline
\multicolumn{6}{|c|}{$EKSet$} \\ \hline
Context Size &$\psi_{G}$ &$\psi_{P}$ &$\psi_{H}$ &$\psi_{C}$ &$ALL$ \\
    \hline  \hline
    1   &44.9\%&   44.9\%&   51.8\%&   52.4\%&   65.8\%    \\ \hline
    2   &57.3\%&   52.8\%&   61.7\%&   64.4\%&   74.4\%    \\ \hline
    3   &58.8\%&   55.2\%&   64.1\%&   65.5\%&   75.8\% \\ \hline
    4   &56.1\%&   54.6\%&   61.8\%&   64.3\%&   74.4\% \\ \hline
    5   &53.7\%&   52.6\%&   60.4\%&   62.5\%&   73.9\%  \\ \hline \hline
\multicolumn{6}{|c|}{$EJSet$} \\ \hline
Context Size &$\psi_{G}$ &$\psi_{P}$ &$\psi_{H}$ &$\psi_{C}$ &$ALL$ \\
    \hline  \hline
    1   &46.4\%&   52.1\%&   58.0\%&   62.0\%&   70.4\%    \\ \hline
    2   &58.2\%&   59.5\%&   65.6\%&   68.7\%&   76.3\%    \\ \hline
    3   &58.8\%&   59.2\%&   65.8\%&   69.1\%&   77.0\% \\ \hline
    4   &56.4\%&   58.5\%&   64.4\%&   68.2\%&   76.0\% \\ \hline
    5   &53.9\%&   56.4\%&   62.9\%&   66.3\%&   75.5\%  \\ \hline
\end{tabular}
\caption{\label{tab10} Results of Context Window-Size Test: ALL
means $\psi_{G+P+H+C}$.}
\end{center}
\end{table}

\subsection{Training Data-Size Test} \label{sect:exp_size}
    Table~\ref{tab11} shows the results of the Training Data-Size Test
using MEM-based machine transliteration models.
    We evaluated the performance of the four models and $ALL$ while varying the size of
the training data from 20\% to 100\%.
    Obviously, the more training
data used, the higher the system performance.
    However, the objective of this test was to determine whether the
transliteration models perform reasonably well even for a small
amount of training data.
    We found that $\psi_{G}$ was the most
sensitive of the four models to the amount of training data; it had
the largest difference in performance between 20\% and 100\%.
    In contrast, \textit{ALL} showed the smallest performance gap.
    The results of this test shows that combining different
transliteration models is helpful in producing correct
transliterations even if there is little training data.

\begin{table}[ht]
\begin{center}
\begin{tabular}{|c|c|c|c|c|c|}
  \hline
\multicolumn{6}{|c|}{$EKSet$} \\ \hline
Training Data Size &$\psi_{G}$ &$\psi_{P}$ &$\psi_{H}$ &$\psi_{C}$ &$ALL$ \\
    \hline  \hline
    20\%   &46.6\%&   47.3\%&   53.4\%&   57.0\%&   67.5\%    \\ \hline
    40\%   &52.6\%&   51.5\%&   58.7\%&   62.1\%&   71.6\%    \\ \hline
    60\%   &55.2\%&   53.0\%&   61.5\%&   63.3\%&   73.0\% \\ \hline
    80\%   &58.9\%&   54.0\%&   62.6\%&   64.6\%&   74.7\% \\ \hline
    100\%   &58.8\%&   55.2\%&   64.1\%&   65.5\%&   75.8\%  \\ \hline \hline
\multicolumn{6}{|c|}{$EJSet$} \\ \hline
Training Data Size &$\psi_{G}$ &$\psi_{P}$ &$\psi_{H}$ &$\psi_{C}$ &$ALL$ \\
    \hline  \hline
    20\%   &47.6\%&   51.2\%&   56.4\%&   60.4\%&   69.6\%    \\ \hline
    40\%   &52.4\%&   55.1\%&   60.7\%&   64.8\%&   72.6\%    \\ \hline
    60\%   &55.2\%&   57.3\%&   62.9\%&   66.6\%&   74.7\% \\ \hline
    80\%   &57.9\%&   58.8\%&   65.4\%&   68.0\%&   76.7\% \\ \hline
    100\%   &58.8\%&   59.2\%&   65.8\%&   69.1\%&   77.0\%  \\ \hline
\end{tabular}
\caption{\label{tab11}Results of Training Data-Size Test: ALL means
$\psi_{G+P+H+C}$.}
\end{center}
\end{table}


\section{Discussion}
\label{sec:discuss}
\begin{figure}[ht]
     \centering
     \subfigure[$\psi_G$+$\psi_P$+$\psi_C$]{
          \label{fig7-1}
          \centering
          \includegraphics[width=0.200\textwidth,clip]{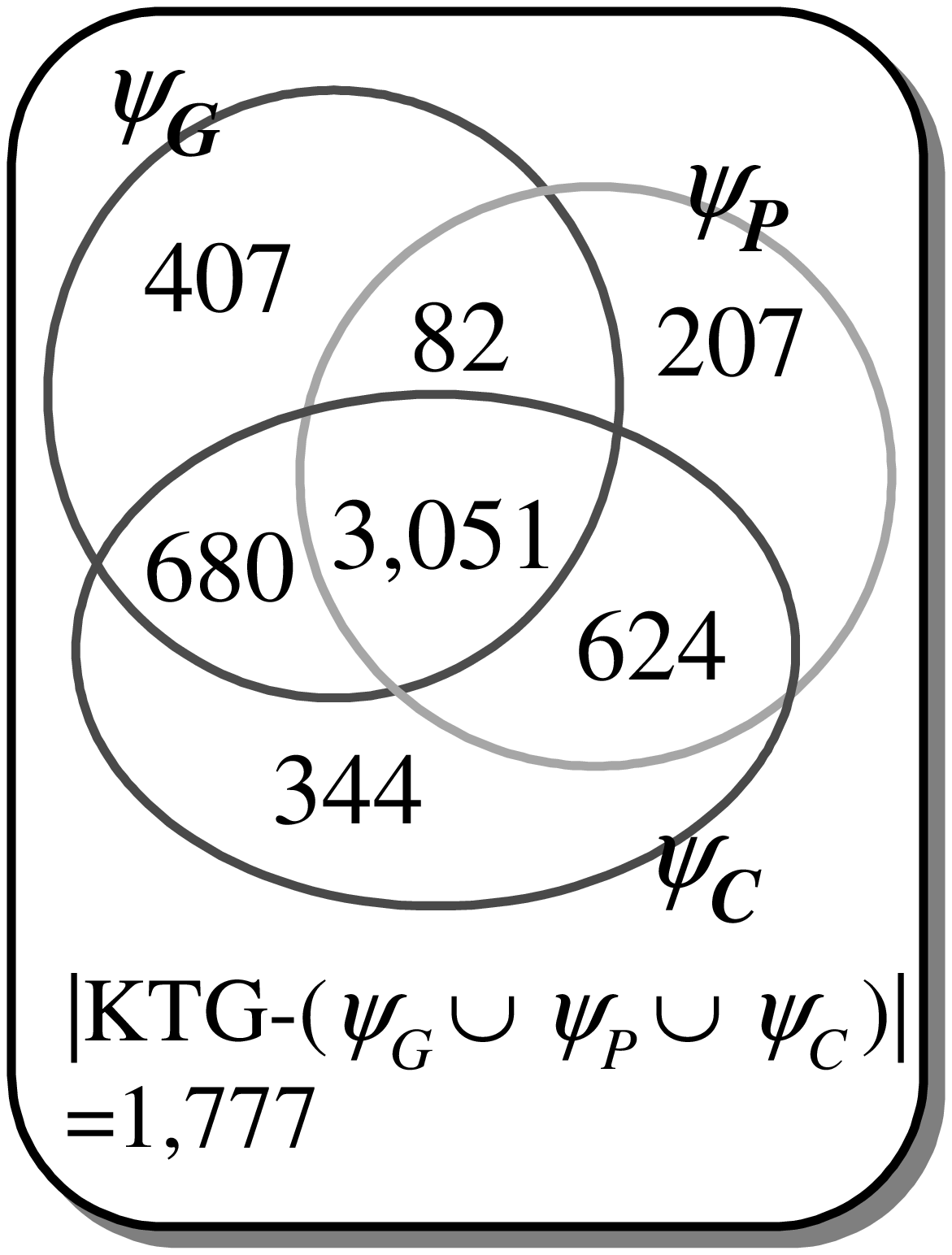}}
     \subfigure[$\psi_G$+$\psi_P$+$\psi_H$]{
          \label{fig7-2}
          \includegraphics[width=0.200\textwidth,clip]{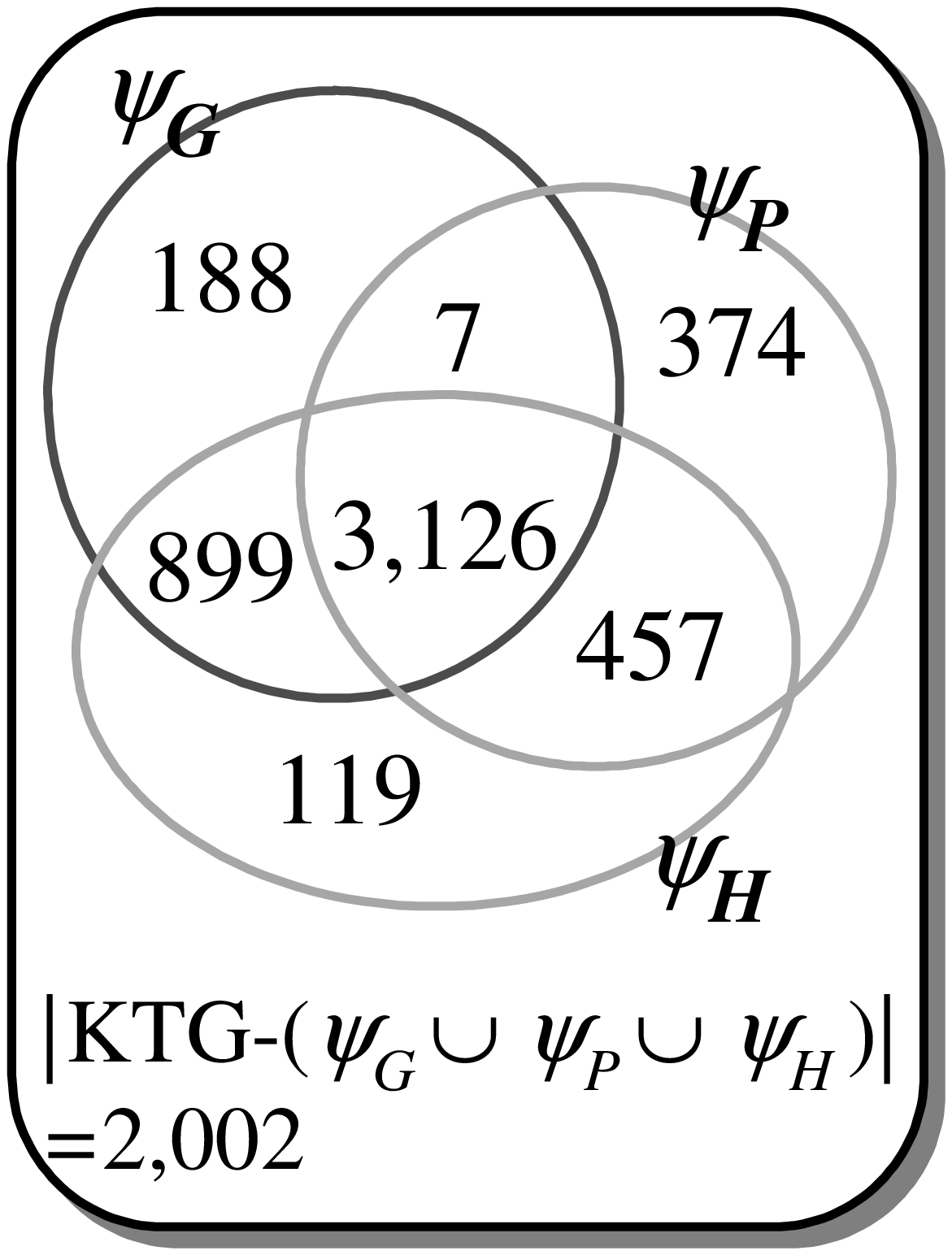}}
     \subfigure[$\psi_P$+$\psi_H$+$\psi_C$]{
           \label{fig7-3}
           \includegraphics[width=0.200\textwidth,clip]{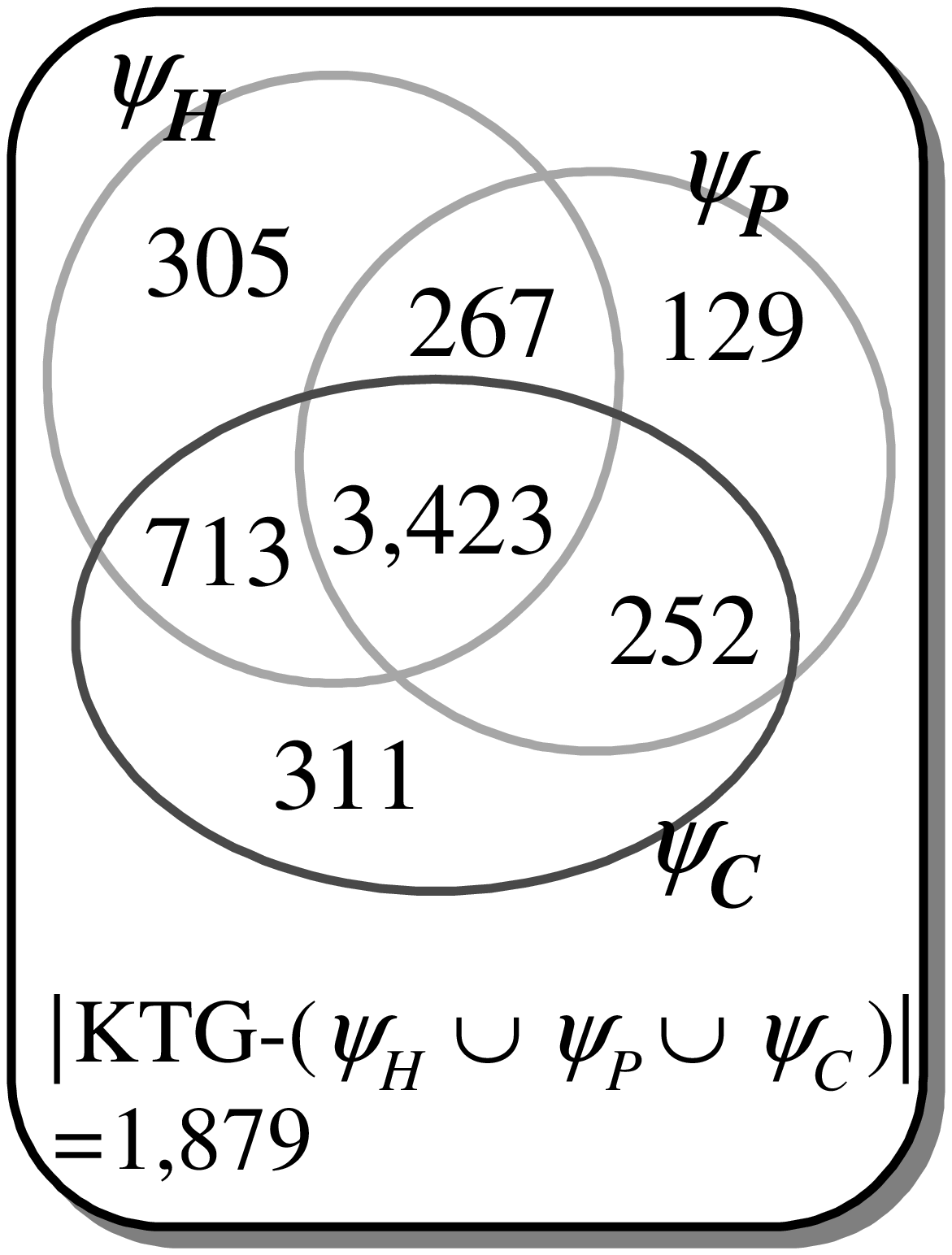}}
     \subfigure[$\psi_G$+$\psi_H$+$\psi_C$]{
           \label{fig7-4}
           \includegraphics[width=0.200\textwidth,clip]{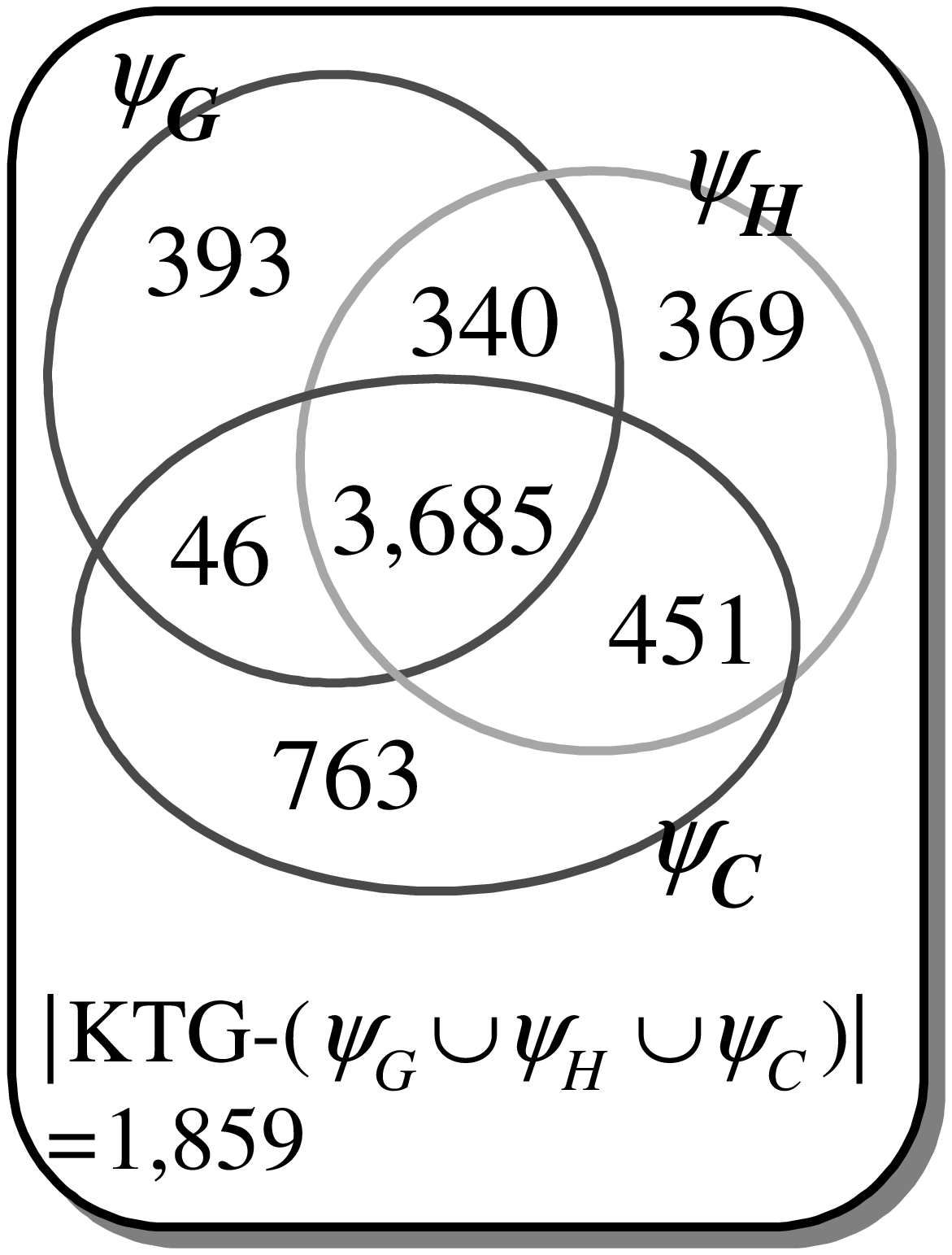}} 
     \caption{Distributions of correct transliterations produced by models for
English-to-Korean transliteration. KTG represents ``Korean
Transliterations in the Gold standard". Note that $|\psi_G$ $\cup$
$\psi_P$ $\cup$ $\psi_H$ $\cup$ $\psi_C|$ = 5,439, $|\psi_G$ $\cap$
$\psi_P$ $\cap$ $\psi_H$ $\cap$ $\psi_C|$ = 3,047, and $|KTG|$ =
7,172.}
     \label{fig7}
\end{figure}

\begin{figure}[ht]
     \centering
     \subfigure[$\psi_G$+$\psi_P$+$\psi_C$]{
          \label{fig8-1}

          \includegraphics[width=0.200\textwidth,clip]{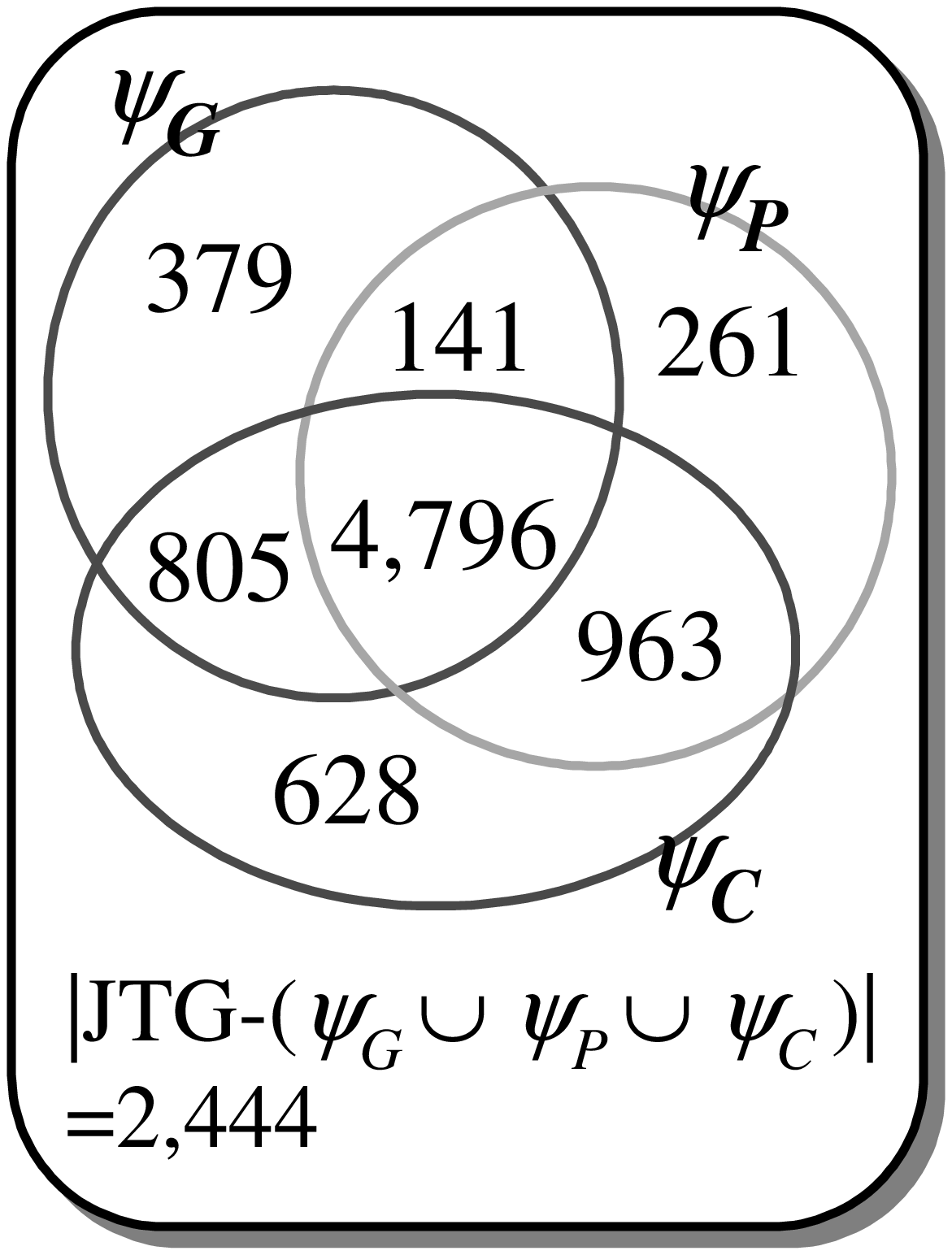}}
     \subfigure[$\psi_G$+$\psi_P$+$\psi_H$]{
          \label{fig8-2}
          \includegraphics[width=0.200\textwidth,clip]{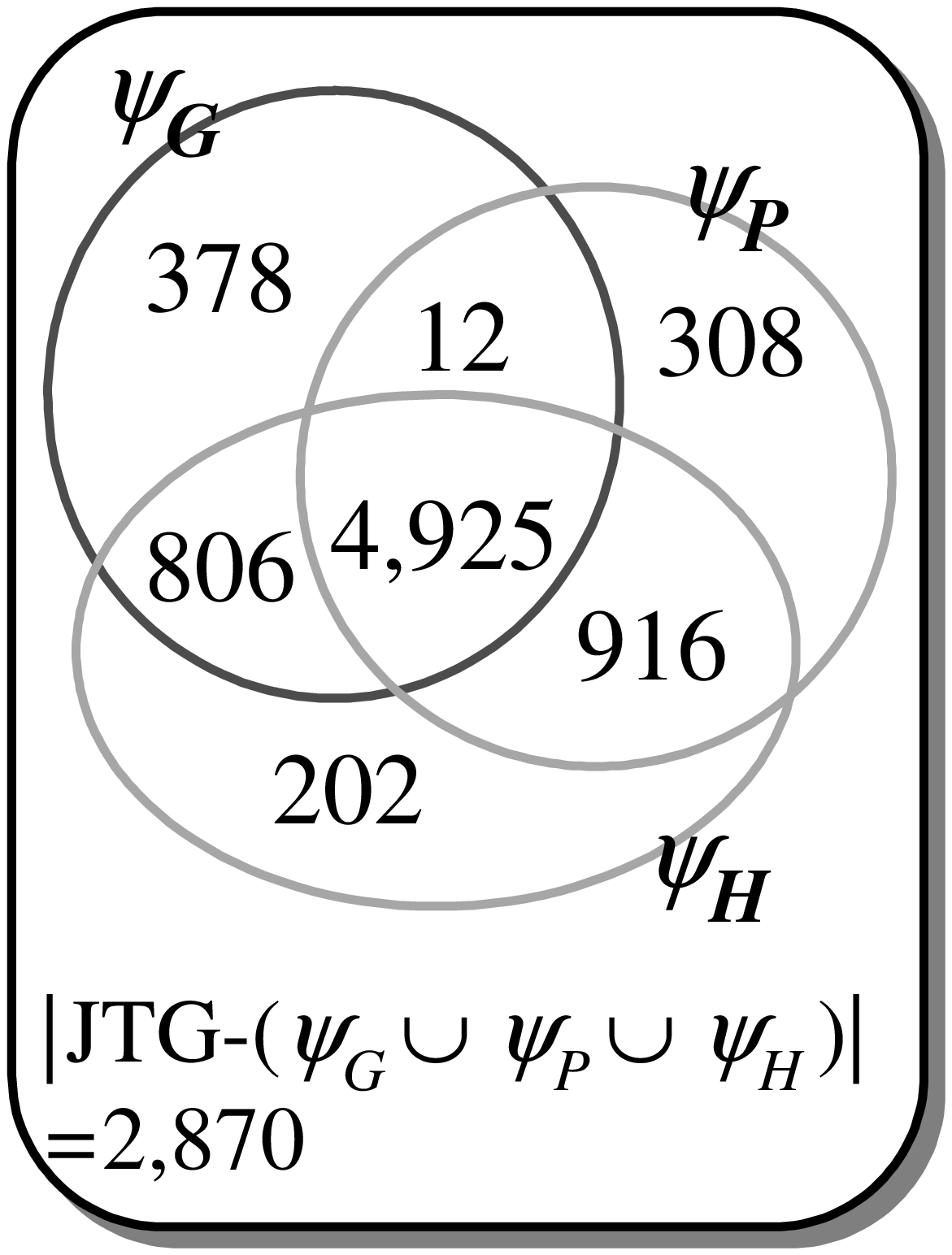}}
     \subfigure[$\psi_P$+$\psi_H$+$\psi_C$]{
           \label{fig8-3}
           \includegraphics[width=0.200\textwidth,clip]{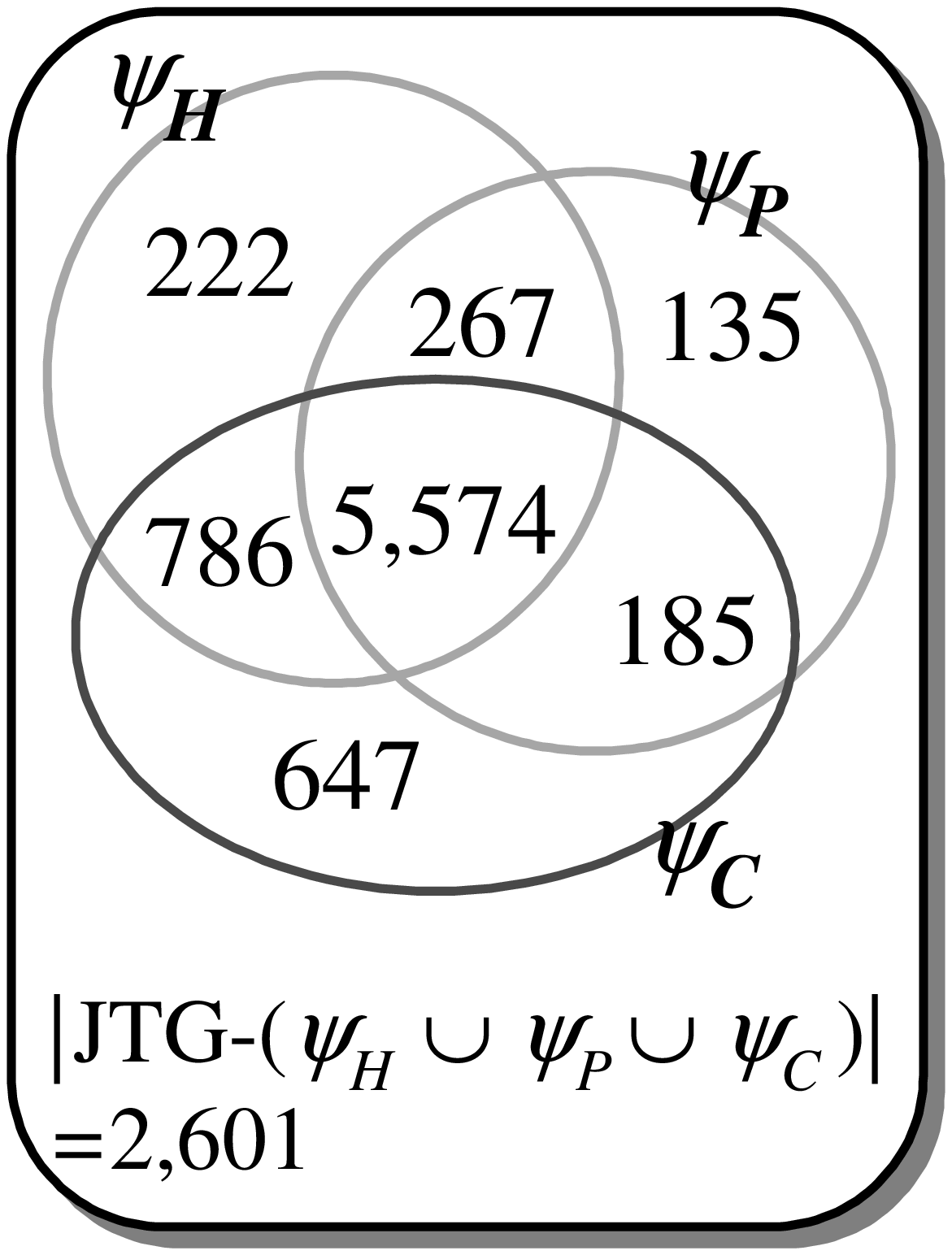}}
     \subfigure[$\psi_G$+$\psi_H$+$\psi_C$]{
           \label{fig8-4}
           \includegraphics[width=0.200\textwidth,clip]{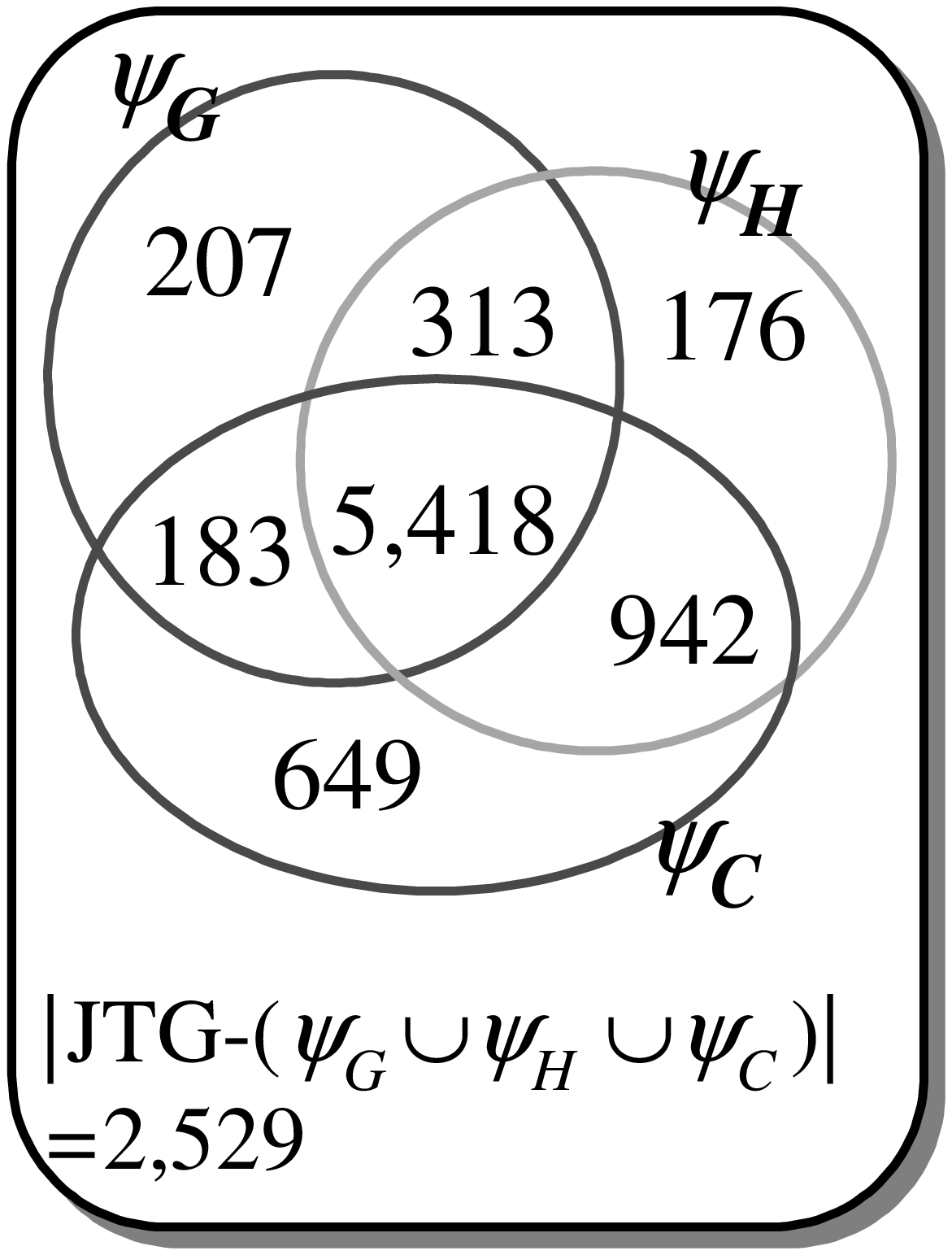}} 
     \caption{Distributions of correct transliterations produced by models
     for English-to-Japanese transliteration. JTG represents
        ``Japanese Transliterations in the Gold standard".
        Note that $|\psi_G$ $\cup$ $\psi_P$ $\cup$  $\psi_H$ $\cup$
        $\psi_C|$=8,021,
        $|\psi_G$ $\cap$ $\psi_P$ $\cap$  $\psi_H$ $\cap$ $\psi_C|$=4,786, and
        $|JTG|$ = 10,417.}
     \label{fig8}
\end{figure}

    Figures~\ref{fig7} and \ref{fig8} show the distribution of the
correct transliterations produced by each transliteration model and
by the combination of models, all based on the MEM.
    The $\psi_G$, $\psi_P$, $\psi_H$, and $\psi_C$ in the figures
represent the set of correct transliterations produced by each model
through $k$-fold validation.
    For example, $|\psi_G|$ = 4,220
for EKSet and $|\psi_G|$ = 6,121 for EJSet mean that $\psi_G$
produced 4,220 correct transliterations for 7,172 English words in
EKSet ($|KTG|$ in Figure~\ref{fig7}) and 6,121 correct ones for
10,417 English words in EJSet ($|JTG|$ in Figure~\ref{fig8}).
    An important factor in modeling a transliteration model is to reflect
the dynamic transliteration behaviors, which means that a
transliteration process dynamically uses the source grapheme and
source phoneme to produce transliterations.
    Due to these dynamic behaviors, a
transliteration can be grapheme-based transliteration, phoneme-based
transliteration, or some combination of the two.
    The forms of transliterations are classified on the basis of the
information upon which the transliteration process mainly relies
(either a source grapheme or a source phoneme or some combination of
the two).
    Therefore, an effective transliteration system should
be able to produce various types of transliterations at the same
time.
    One way to accommodate the different dynamic transliteration
behaviors is to combine different transliteration models, each of
which can handle a different behavior.
    Synergy can be achieved by
combining models so that one model can produce the correct
transliteration when the others cannot.
    Naturally, if the models
tend to produce the same transliteration, less synergy can be
realized from combining them.
    Figures~\ref{fig7} and \ref{fig8}
show the synergy gained from combining transliteration models in
terms of the size of the intersection and the union of the
transliteration models.

    The figures show that, as the area of intersection between
different transliteration models becomes smaller, the size of their
union tends to become bigger.
    The main characteristics
obtained from these figures are summarized in Table~\ref{tab10-1}.
\begin{table}[ht]
\begin{center}
\begin{tabular}{ |c|c|c| }
  \hline
     &EKSet & EJSet \\ \hline \hline
         $|\psi_G|$ &4,202 & 6,118 \\ \hline
         $|\psi_P|$ &3,947 & 6,158 \\ \hline
         $|\psi_H|$ &4,583 & 6,846 \\ \hline
         $|\psi_C|$ &4,680 & 7,189 \\ \hline
         $|\psi_G \cap \psi_P|$ &3,133 & 4,937 \\ \hline
         $|\psi_G \cap \psi_C|$ &3,731 & 5,601 \\ \hline
         $|\psi_G \cap \psi_H|$ &4,025 & 5,731 \\ \hline
         $|\psi_C \cap \psi_H|$ &4,136 & 6,360 \\ \hline
         $|\psi_P \cap \psi_C|$ &3,675 & 5,759 \\ \hline
         $|\psi_P \cap \psi_H|$ &3,583 & 5,841 \\ \hline
         $|\psi_G \cup \psi_P|$ &5,051 & 7,345 \\ \hline
         $|\psi_G \cup \psi_C|$ &5,188 & 7,712 \\ \hline
         $|\psi_G \cup \psi_H|$ &4,796 & 7,239 \\ \hline
         $|\psi_C \cup \psi_H|$ &5,164 & 7,681 \\ \hline
         $|\psi_P \cup \psi_C|$ &4,988 & 7,594 \\ \hline
         $|\psi_P \cup \psi_H|$ &4,982 & 7,169 \\ \hline
\end{tabular}
\caption{\label{tab10-1} Main characteristics obtained from
Figures~\ref{fig7} and \ref{fig8}.}
\end{center}
\end{table}
    The first thing to note is that $|\psi_G \cap \psi_P|$ is clearly
smaller than any other intersection.
    The main reason for this is
that $\psi_G$ and $\psi_P$ use no common information ($\psi_G$ uses
source graphemes while $\psi_P$ uses source phonemes).
    However, the others use at least one of source grapheme and source
phoneme (source graphemes are information common to $\psi_G$,
$\psi_H$, and $\psi_C$ while source phonemes are information common
to $\psi_P$, $\psi_H$, and $\psi_C$).
    Therefore, we can infer that the synergy derived from combining $\psi_G$ and
$\psi_P$ is greater than that derived from the other combinations.
    However, the size of the union between the various pairs of
transliteration models in Table~\ref{tab10-1} shows that $|\psi_C
\cup \psi_H|$ and $|\psi_G \cup \psi_C|$ are bigger than $|\psi_G
\cup \psi_P|$.
    The main reason for this might be the higher
transliteration power of $\psi_C$ and $\psi_H$ compared to that of
$\psi_G$ and $\psi_P$ -- $\psi_C$ and $\psi_H$ cover more of the KTG
and JTG than both $\psi_G$ and $\psi_P$.
    The second thing to note is that the contribution of each transliteration model to
$|\psi_G$ $\cup$ $\psi_P$ $\cup$ $\psi_H$ $\cup$ $\psi_C|$ can be
estimated from the difference between $|\psi_G$ $\cup$ $\psi_P$
$\cup$ $\psi_H$ $\cup$ $\psi_C|$ and the union of the three other
transliteration models.
    For example, we can measure the
contribution of $\psi_H$ from the difference between $|\psi_G$
$\cup$ $\psi_P$ $\cup$ $\psi_H$ $\cup$ $\psi_C|$ and $|\psi_G$
$\cup$ $\psi_P$ $\cup$ $\psi_C|$.
    As shown in Figures~\ref{fig7-1} and \ref{fig8-1}),
$\psi_H$ makes the smallest contribution while $\psi_C$
(Figures~\ref{fig7-2} and \ref{fig8-2}) makes the largest
contribution.
    The main reason for $\psi_H$ making the smallest
contribution is that it tends to produce the same transliteration as
the others, so the intersection between $\psi_H$ and the others
tends to be large.

    It is also important to rank the transliterations produced by a
transliteration system for a source language word on the basis of
their relevance.
    While a transliteration system can produce a list
of transliterations, each reflecting a dynamic transliteration
behavior, it will fail to perform well unless it can distinguish
between correct and wrong transliterations.
    Therefore, a transliteration system should be able to produce various kinds of
transliterations depending on the dynamic transliteration behaviors
and be able to rank them on the basis of their relevance.
    In addition, the application of transliteration results
to natural language applications such as machine translation and
information retrieval requires that the transliterations be ranked
and assigned a relevance score.

    In summary, 1) \textbf{producing a list of transliterations
reflecting dynamic transliteration behaviors} (one way is to combine
the results of different transliteration models, each reflecting one
of the dynamic transliteration behaviors) and 2) \textbf{ranking the
transliterations in terms of their relevance} are both necessary to
improve the performance of machine transliteration.
    In the next section, we describe a way to
calculate the relevance of transliterations produced by a
combination of the four transliteration models.


\section{Transliteration Ranking}
\label{sec:ranking}
    The basic assumption of transliteration ranking is that correct
transliterations are more frequently used in real-world texts than
incorrect ones.
    Web data reflecting the real-world usage of
transliterations can thus be used as a language resource to rank
transliterations.
    Transliterations that appear more frequently in
web documents are given either a higher rank or a higher score.
    The goal of transliteration ranking, therefore, is to rank correct
transliterations higher and rank incorrect ones lower.
    The transliterations produced for a given English word by the four
transliteration models ($\psi_G$, $\psi_P$, $\psi_H$, and $\psi_C$),
based on the MEM, were ranked using web data.

\begin{figure}[ht]
\begin{center}
\includegraphics[width=0.800\textwidth]{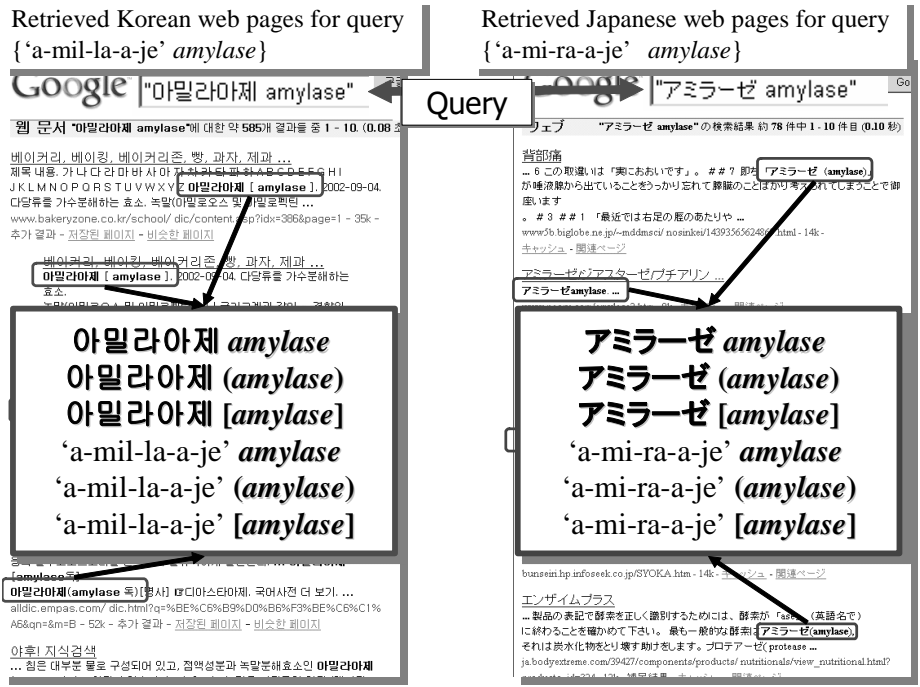}
\end{center}
\caption{\label{fig9} Desirable retrieved web pages for
transliteration ranking.}
\end{figure}

    Our transliteration ranking relies on web frequency (number of web
documents).
    To obtain reliable web frequencies, it is important to
consider a transliteration and its corresponding source language
word together rather than the transliteration alone.
    This is because our aim is to find correct transliterations corresponding
to a source language word rather than to find transliterations that
are frequently used in the target language.
    Therefore, the best approach to transliteration ranking using web data is
to find web documents in which transliterations are used as
translations of the source language word.

    A bilingual phrasal search (BPS) retrieves the Web with a Web search engine query,
which is a phrase composed of a transliteration and its source
language word (e.g., \{`a-mil-la-a-je' \textit{amylase}\}).
    The BPS enables the Web search engine to find web documents that contain correct
transliterations corresponding to the source language word.
    Note that a phrasal query is represented in brackets, where
the first part is a transliteration and the second part is the
corresponding source language word.
    Figure~\ref{fig9} shows Korean
and Japanese web documents retrieved using a BPS for
\textit{amylase} and its Korean/Japanese transliterations,
`a-mil-la-a-je' and `a-mi-ra-a-je'.
    The web documents retrieved by
a BPS usually contain a transliteration and its corresponding source
language word as a translation pair, with one of them often placed
in parentheses, as shown in Figure~\ref{fig9}.

    A dilemma arises, though, regarding the quality and coverage of
retrieved web documents.
    Though a BPS generally provides
high-quality web documents that contain correct transliterations
corresponding to the source language word, the coverage is
relatively low, meaning that it may not find any web documents for
some transliterations.
    For example, a BPS for the Japanese phrasal
query \{`a-ru-ka-ro-si-su' \textit{alkalosis}\} and the Korean
phrasal query \{`eo-min' \textit{ermine}\} found no web documents.
    Therefore, alternative search methods are necessary when the BPS
fails to find any relevant web documents.
    A bilingual keyword search (BKS)~\cite{qu04,huang05,zhang05} can be used when the BPS
fails, and a monolingual keyword search (MKS)~\cite{grefenstette04}
can be used when both the BPS and BKS fail.
    Like a BPS, a BKS makes use of two keywords, a
transliteration and its source language word, as a search engine
query.
    Whereas a BPS retrieves web documents containing the two
keywords as a phrase, a BKS retrieves web documents containing them
anywhere in the document.
    This means that the web frequencies
of noisy transliterations are sometimes higher than those of correct
transliterations in a BKS, especially when the noisy
transliterations are one-syllable transliterations.
    For example, `mok', which is a Korean transliteration produced for
\textit{mook} and a one-syllable noisy transliteration, has a higher
web frequency than `mu-keu', which is the correct transliteration
for \textit{mook}, because `mok' is a common Korean noun that
frequently appears in Korean texts with the meaning of
\textit{neck}.
    However, a BKS can improve coverage
without a great loss of quality in the retrieved web documents if
the transliterations are composed of two or more syllables.

    Though a BKS has higher coverage than a BPS, it can fail to
retrieve web documents in some cases.
    In such cases, an MKS~\cite{grefenstette04} is used.
    In an MKS, a transliteration alone is used as the search engine query.
    A BPS and a BKS act like a translation model, while an MKS acts like a language model.
    Though an MKS cannot give information as to whether
the transliteration is correct, it does provide information as to
whether the transliteration is likely to be a target language word.
The three search methods are used sequentially (BPS, BKS, MKS).
    If one method fails to retrieve any relevant web documents,
the next one is used. Table~\ref{tab12} summarizes the conditions
for applying each search method.

    Along with these three search strategies, three different search
engines are used to obtain more web documents. The search engines
used for this purpose should satisfy two conditions: 1) support
Korean/Japanese web document retrieval and 2) support both phrasal
and keyword searches.
    Google\footnote{\url{http://www.google.com}},
Yahoo\footnote{\url{http://www.yahoo.com}}, and
MSN\footnote{\url{http://www.msn.com}} satisfy these conditions, and
we used them as our search engines.

\begin{table}[ht]
\begin{center}
\begin{tabular}{|c|c|}
  \hline
Search method & Condition \\ \hline \hline
BPS & $\sum_j \sum_{c_k \in C} WF_{BPSj}(e,c_k)) \ne 0 $ \\
 \hline
  \multirow{2}{*}{BKS} & $\sum_j \sum_{c_k \in C} WF_{BPSj}(e,c_k)) = 0 $ \\

     & $\sum_j \sum_{c_k \in C} WF_{BKSj}(e,c_k)) \ne 0 $  \\
        \hline
  \multirow{3}{*}{MKS} & $\sum_j \sum_{c_k \in C} WF_{BPSj}(e,c_k) = 0$  \\
     & $\sum_j \sum_{c_k \in C} WF_{BKSj}(e,c_k) = 0$ \\
     & $\sum_j \sum_{c_k \in C} WF_{MKSj}(e,c_k) \ne 0$ \\ \hline
\end{tabular}
\caption{\label{tab12} Conditions under which each search method is
applied.}
\end{center}
\end{table}

\begin{eqnarray}
\label{eqn5}
    RF(e,c_i) = \sum_j NWF_{j}(e,c_i)
    = \sum_j \frac{WF_{j}(e,c_i)}{\sum_{c_k \in C} WF_j(e,c_k)}
\end{eqnarray}

    Web frequencies acquired from these three search methods and these
three search engines were used to rank transliterations on the basis
of Formula~(\ref{eqn5}), where $c_i$ is the $i^{th}$ transliteration
produced by the four transliteration models, $e$ is the source
language word of $c_i$, $RF$ is a function for ranking
transliterations, $WF$ is a function for calculating web frequency,
$NWF$ is a function for normalizing web frequency, $C$ is a set of
produced transliterations, and $j$ is an index for the $j^{th}$
search engine.
    We used the normalized web frequency as a
ranking factor.
    The normalized web frequency is the web frequency
divided by the total web frequency of all produced transliterations
corresponding to one source language word.
    The score for a transliteration is then calculated by summing up
the normalized web frequencies of the transliteration given by the
three search engines.
    Table~\ref{tab13} shows an example ranking
for the English word \textit{data} and its possible Korean
transliterations, `de-i-teo', `de-i-ta', and `de-ta', which web
frequencies are obtained using a BPS.
    The normalized $WF_{BPS}$ ($NWF_{BPS}$) for search
engine \textsf{A} was calculated as follows.

\begin{itemize}
    \item $NWF_{BPS}$ (\textit{data}, `de-i-teo') = 94,100 / (94,100 + 67,800 + 54) = 0.5811
    \item $NWF_{BPS}$ (\textit{data}, `de-i-ta') = 67,800 / (94,100 + 67,800 + 54) = 0.4186
    \item $NWF_{BPS}$ (\textit{data}, `de-ta') = 54 / (94,100 + 67,800 + 54) = 0.0003
\end{itemize}
    The ranking score for `de-i-teo' was then calculated by summing up
$NWF_{BPS}$ (\textit{data}, `de-i-teo') for each search engine:
\begin{itemize}
    \item $RF_{BPS}$ (\textit{data}, `de-i-teo') = 0.5810 + 0.7957 + 0.3080 = 1.6848
\end{itemize}

\begin{table}[ht]
\begin{center}
\begin{tabular}{|c|c|c|c|c|c|c|}
  \hline
\multirow{3}{*}{Search Engine}  &
\multicolumn{6}{|c|}{\textit{e=data}}
\\ \cline{2-7}
  & \multicolumn{2}{|c|}{$c_1$= `de-i-teo'}&
\multicolumn{2}{|c|}{$c_2$= `de-i-ta'} & \multicolumn{2}{|c|}{$c_3$=
    `de-ta'}  \\ \cline{2-7}
    &$WF$   &$NWF$  &$WF$   &$NWF$  &$WF$  &
    $NWF$  \\ \hline
 A   &94,100   &0.5811  &67,800   &0.4186  &54  &0.0003   \\ \hline
 B   &101,834  &0.7957  &26,132   &0.2042  &11  &0.0001   \\ \hline
 C   &1,358    &0.3080  &3,028    &0.6868  &23  &0.0052     \\ \hline
 $RF$&  \multicolumn{2}{|c|}{1.6848}&\multicolumn{2}{|c|}{1.3096} &
 \multicolumn{2}{|c|}{0.0056} \\ \hline
\end{tabular}
\caption{\label{tab13} Example transliteration ranking for
\textit{data} and its transliterations; $WF$, $NWF$, and $RF$
represent $WF_{BPS}$, $NWF_{BPS}$, and $RF_{BPS}$, respectively.}
\end{center}
\end{table}

\subsection{Evaluation}
    We tested the performance of the transliteration ranking under two
conditions: 1) with all test data (ALL) and 2) with test data for
which at least one transliteration model produced the correct
transliteration (CTC).
    Testing with ALL showed the overall
performance of the machine transliteration while testing with CTC
showed the performance of the transliteration ranking alone.
    We used the performance of the individual transliteration models
($\psi_G$, $\psi_P$, $\psi_H$, and $\psi_C$) as the baseline.
    The results are shown in Table~\ref{tab14}.
    ``Top-n" means that the correct transliteration was within the Top-n ranked
transliterations.
    The average number of produced Korean transliterations was 3.87 and
that of Japanese ones was 4.50; note that $\psi_P$ and $\psi_C$
produced more than one transliteration because of pronunciation
variations.
    The results for both English-to-Korean and
English-to-Japanese transliteration indicate that our ranking method
effectively filters out noisy transliterations and positions the
correct transliterations in the top rank; most of the correct
transliterations were in Top-1.
    We see that
transliteration ranking (in Top-1) significantly improved
performance of the individual models for both EKSet and
EJSet\footnote{A one-tail paired t-test showed that the performance
improvement was significant (level of significance = 0.001.}.
    The overall performance of the
transliteration (for ALL) as well that of the ranking (for CTC) were
relatively good.
    Notably, the CTC performance showed that web
data is a useful language resource for ranking transliterations.

\begin{table}[ht]
\begin{center}
\begin{tabular}{|c|c|c|c|}
  \hline
Test data &    & EKSet  & EJSet \\ \hline \hline
    \multirow{4}{*}{ALL} &$\psi_G$   &58.8\% &58.8\% \\ \cline{2-4}
        &$\psi_P$   &55.2\%   &59.2\% \\ \cline{2-4}
        &$\psi_H$   &64.1\%   &65.8\% \\ \cline{2-4}
        &$\psi_C$   &65.5\%   &69.1\% \\ \hline
    \multirow{3}{*}{ALL} &Top-1   &71.5\% &74.8\% \\ \cline{2-4}
        &Top-2   &75.3\%   &76.9\% \\ \cline{2-4}
        &Top-3   &75.8\%   &77.0\% \\ \hline
    \multirow{3}{*}{CTC} &Top-1   &94.3\%  &97.2\% \\ \cline{2-4}
        &Top-2   &99.2\%    &99.9\% \\ \cline{2-4}
        &Top-3   &100\%    &100\% \\ \hline
\end{tabular}
\caption{\label{tab14} Results of Transliteration ranking.}
\end{center}
\end{table}

\subsection{Analysis of Results}
    We defined two error types: \textbf{production errors} and
\textbf{ranking errors}.
    A production error is when there is no
correct transliteration among the produced transliterations.
    A ranking error is when the correct transliteration does not appear
in the Top-1 ranked transliterations.

    We examined the relationship
between the search method and the transliteration ranking.
    Table~\ref{tab15} shows the ranking performance by each search method.
    The RTC represents correct transliterations
ranked by each search method.
    The NTC represents test data ranked,
that is, the coverage of each search method.
    The ratio of RTC to NTC represents the upper bound of
performance and the difference between RTC and NTC is the number of
errors.

    The best performance was with a BPS.
    A BPS handled 5,270 out of 7,172 cases for EKSet and 8,829 out of 10,417 cases
for EJSet.
    That is, it did the best job of retrieving web
documents containing transliteration pairs.
    Analysis of the ranking errors revealed that the main cause of such errors in a
BPS was transliteration variations.
    These variations contribute to ranking errors in two ways.
    First, when the web frequencies of transliteration variations are
higher than those of the standard ones, the variations are ranked
higher than the standard ones, as shown by the examples in
Table~\ref{tab16}.
    Second, when the transliterations include only
transliteration variations (i.e., there are no correct
transliterations), the correct ranking cannot be.
    In this case, ranking errors are caused by production errors.
    With a BPS, there were 603 cases of this for EKSet and 895 cases for EJSet.

\begin{table}[ht]
\begin{center}
\begin{tabular}{|c|c|c|c|c|c|c|}
  \hline
    &\multicolumn{3}{|c|}{EKSet}  & \multicolumn{3}{|c|}{EJSet} \\
    \cline{2-7}
    &BPS &BKS &MKS &BPS &BKS &MKS \\ \hline \hline
    Top-1 &  83.8\%  &55.1\%  &16.7\%  &86.2\%  &19.0\% &2.7\% \\
    \hline
    Top-2 &  86.6\%  &58.4\%  &27.0\%  &88.3\%  &22.8\% &4.2\% \\
    \hline
    Top-3 & 86.6\%  &58.2\%  &31.3\%  &88.35\%  &22.9\%  &4.3\% \\
    \hline
    RTC & 4,568 &596 & 275 &7,800 &188  &33 \\ \hline
    NTC & 5,270 &1,024  &878 &8,829 &820  &768 \\ \hline
\end{tabular}
\caption{\label{tab15} Ranking performance of each search method.}
\end{center}
\end{table}

\begin{table}[ht]
\begin{center}
\begin{tabular}{|l|l|l|}
  \hline
                                    &  Transliteration &   Web Frequency \\
                                    \hline \hline
    \multirow{2}{*}{compact $\rightarrow$ Korean}    & `\textsf{kom-paek-teu}'  & 1,075 \\ \cline{2-3}
                                    & `keom-paek-teu'$\ast$       & 1,793  \\ \hline
    \multirow{2}{*}{pathos $\rightarrow$ Korean}     & `\textsf{pa-to-seu}'     & 1,615 \\ \cline{2-3}
                                    & `pae-to-seu'$\ast$          & 14,062  \\ \hline
    \multirow{2}{*}{cohen $\rightarrow$ Japanese}    & `\textsf{ko-o-he-n}'     & 23 \\ \cline{2-3}
                                    & `ko-o-e-n'$\ast$            & 112  \\ \hline
    \multirow{2}{*}{criteria $\rightarrow$ Japanese} & `\textsf{ku-ra-i-te-ri-a}' & 104 \\ \cline{2-3}
                                    & `ku-ri-te-ri-a'$\ast$       & 1,050  \\ \hline
\end{tabular}
\caption{\label{tab16} Example ranking errors when a BPS was used
($\ast$ indicates a variation).}
\end{center}
\end{table}

    NTC was smaller with a BKS and an MKS because a BPS retrieves web
documents whenever possible.
    Table~\ref{tab15} shows that production errors are the main reason
a BPS fails to retrieve web documents. (When a BKS or MKS was used,
production errors occurred in all but 871\footnote{596 (RTC of BKS
in EKSet) + 275 (RTC of MKS in EKSet) = 871} cases for EKSet and
221\footnote{188 (RTC of BKS for EJSet) + 33 (RTC of MKS for EJSet)
= 221} cases for EJSet).
    The results also show that a BKS was more effective than
an MKS.

    The trade-off between the quality and coverage of retrieved web
documents is an important factor in transliteration ranking.
    A BPS provides better quality rather than wider coverage, but is
effective since it provides reasonable coverage.


\section{Conclusion}
\label{sec:conclusion}
    We tested and compared four transliteration models, \textbf{grapheme-based
transliteration model} ($\psi_G$), \textbf{phoneme-based
transliteration model} ($\psi_P$), \textbf{hybrid transliteration
model} ($\psi_H$), and \textbf{correspondence-based transliteration
model }($\psi_C$), for English-to-Korean and English-to-Japanese
transliteration.
    We modeled a framework for
the four transliteration models and compared them within the
framework.
    Using the results, we examined a way to improve the
performance of machine transliteration.

    We found that the $\psi_H$ and $\psi_C$ are more effective
than the $\psi_G$ and $\psi_P$.
    The main reason for the better
performance of $\psi_C$ is that it uses the correspondence between
the source grapheme and the source phoneme.
    The use of this correspondence positively affected transliteration performance in
various tests.

    We demonstrated that $\psi_G$, $\psi_P$, $\psi_H$, and $\psi_C$
can be used as complementary transliteration models to improve the
chances of producing correct transliterations.
    A combination of the four models produced more correct transliterations both in
English-to-Korean transliteration and English-to-Japanese
transliteration compared to each model alone.
    Given these results,
we described a way to improve machine transliteration that combines
different transliteration models: 1) \textbf{produce a list of
transliterations by combining transliterations produced by multiple
transliteration models}; 2) \textbf{rank the transliterations on the
basis of their relevance}.

    Testing showed that transliteration ranking based on
web frequency is an effective way to calculate the relevance of
transliterations.
    This is because web data reflects real-world usage, so it can be
used to filter out noisy transliterations, which are not used as
target language words or are incorrect transliterations for a source
language word.

    There are several directions for future work.
    Although we considered some transliteration variations,
our test sets mainly covered standard transliterations.
    In corpora or web pages, however, we routinely find other types of
transliterations -- misspelled transliterations, transliterations of
common phrases, etc. -- along with the standard transliterations and
transliteration variations.
    Therefore, further testing using such transliterations is needed to enable
the transliteration models to be compared more precisely.
    To achieve a machine transliteration system capable of higher
performance, we need a more sophisticated transliteration method and
a more sophisticated ranking algorithm.
    Though many correct transliterations can be acquired
through the combination of the four transliteration models, there
are still some transliterations that none of the models can produce.
    We need to devise a method that can produce them.
    Our transliteration ranking method works well, but, because it depends
on web data, it faces limitations if the correct transliteration
does not appear in web data.
    We need a complementary ranking
method to handle such cases.
    Moreover, to demonstrate the
effectiveness of these four transliteration models, we need to apply
them to various natural language processing applications.


\acks{We are grateful to Claire Cardie and the anonymous reviewers
for providing constructive and insightful comments to earlier
drafts of this paper.}



\vskip 0.2in
\bibliographystyle{theapa}
\bibliography{reference}

\end{document}